# WHY FAIRNESS CANNOT BE AUTOMATED: BRIDGING THE GAP BETWEEN EU NON-DISCRIMINATION LAW AND AI

*Sandra Wachter,[1] Brent Mittelstadt,[2] & Chris Russell[3]*

## ABSTRACT

In recent years a substantial literature has emerged concerning bias, discrimination, and fairness in AI and machine learning. Connecting this work to existing legal non-discrimination frameworks is essential to create tools and methods that are practically useful across divergent legal regimes. While much work has been undertaken from an American legal perspective, comparatively little has mapped the effects and requirements of EU law. This Article addresses this critical gap between legal, technical, and organisational notions of algorithmic fairness. Through analysis of EU non-discrimination law and jurisprudence of the European Court of Justice (ECJ) and national courts, we identify a critical incompatibility between European notions of discrimination and existing work on algorithmic and automated fairness. A clear gap exists between statistical measures of fairness as embedded in myriad fairness toolkits and governance mechanisms and the context-sensitive, often intuitive and ambiguous discrimination metrics and evidential requirements used by the ECJ; we refer to this approach as "contextual equality."

This Article makes three contributions. First, we review the evidential requirements to bring a claim under EU non-discrimination law. Due to the disparate nature of algorithmic and human discrimination, the EU's current requirements are too contextual, reliant on intuition, and open to judicial interpretation to be automated. Many of the concepts fundamental to bringing a claim, such as the composition of the disadvantaged and advantaged group, the severity and type of harm suffered, and requirements for the relevance and admissibility of evidence, require normative or political choices to be made by the judiciary on a case-by-case basis. We show that automating fairness or non-discrimination in Europe may be impossible because the law, by design, does not provide a static or homogenous framework suited to testing for discrimination in AI systems.

[1] Oxford Internet Institute, University of Oxford, 1 St. Giles, Oxford, OX1 3JS, UK and The Alan Turing Institute, British Library, 96 Euston Road, London, NW1 2DB, UK. Email: sandra.wachter@oii.ox.ac.uk.

[2] Oxford Internet Institute, University of Oxford, 1 St. Giles, Oxford, OX1 3JS, UK, The Alan Turing Institute, British Library, 96 Euston Road, London, NW1 2DB, UK.

[3] The Alan Turing Institute, British Library, 96 Euston Road, London, NW1 2DB, UK, Department of Electrical and Electronic Engineering, University of Surrey, Guildford, GU2 7HX, UK. This work has been supported by the British Academy, Omidyar Group, Miami Foundation, and the EPSRC via the Alan Turing Institute.

Second, we show how the legal protection offered by non-discrimination law is challenged when AI, not humans, discriminate. Humans discriminate due to negative attitudes (e.g. stereotypes, prejudice) and unintentional biases (e.g. organisational practices or internalised stereotypes) which can act as a signal to victims that discrimination has occurred. Equivalent signalling mechanisms and agency do not exist in algorithmic systems. Compared to traditional forms of discrimination, automated discrimination is more abstract and unintuitive, subtle, intangible, and difficult to detect. The increasing use of algorithms disrupts traditional legal remedies and procedures for detection, investigation, prevention, and correction of discrimination which have predominantly relied upon intuition. Consistent assessment procedures that define a common standard for statistical evidence to detect and assess *prima facie* automated discrimination are urgently needed to support judges, regulators, system controllers and developers, and claimants.

Finally, we examine how existing work on fairness in machine learning lines up with procedures for assessing cases under EU non-discrimination law. A 'gold standard' for assessment of prima facie discrimination has been advanced by the European Court of Justice but not yet translated into standard assessment procedures for automated discrimination. We propose 'conditional demographic disparity' (CDD) as a standard baseline statistical measurement that aligns with the Court's 'gold standard'. Establishing a standard set of statistical evidence for automated discrimination cases can help ensure consistent procedures for assessment, but not judicial interpretation, of cases involving AI and automated systems. Through this proposal for procedural regularity in the identification and assessment of automated discrimination, we clarify how to build considerations of fairness into automated systems as far as possible while still respecting and enabling the contextual approach to judicial interpretation practiced under EU non-discrimination law.

# TABLE OF CONTENTS

## I.   INTRODUCTION

Fairness and discrimination in algorithmic systems are globally recognised as topics of critical importance.[4] To date, a majority of work has started from an American regulatory perspective defined by the notions of 'disparate treatment' and 'disparate impact'.[5] European legal notions of discrimination are not, however, equivalent. In this paper, we examine EU law and jurisprudence of the European Court of Justice concerning non-discrimination. We identify a critical incompatibility between European notions of discrimination and existing work on algorithmic and automated fairness. A clear gap exists between statistical measures of fairness and the context-sensitive, often intuitive and ambiguous discrimination metrics and evidential requirements used by the Court.

---

[4] To name only a few CATHY O'NEIL, WEAPONS OF MATH DESTRUCTION: HOW BIG DATA INCREASES INEQUALITY AND THREATENS DEMOCRACY (2017); Cass R. Sunstein, *Algorithms, correcting biases*, 86 SOC. RES. INT. Q. 499–511 (2019); JOSHUA A. KROLL ET AL., *Accountable Algorithms* (2016), http://papers.ssrn.com/abstract=2765268 (last visited Apr 29, 2016); VIRGINIA EUBANKS, AUTOMATING INEQUALITY: HOW HIGH-TECH TOOLS PROFILE, POLICE, AND PUNISH THE POOR (2018); Katherine J. Strandburg, *Rulemaking and Inscrutable Automated Decision Tools*, 119 COLUMBIA LAW REV. 1851–1886 (2019); MIREILLE HILDEBRANDT & SERGE GUTWIRTH, PROFILING THE EUROPEAN CITIZEN (2008); Mireille Hildebrandt, *Profiling and the Rule of Law*, 1 IDENTITY INF. SOC. IDIS (2009), https://papers.ssrn.com/abstract=1332076 (last visited Jul 31, 2018); Tal Zarsky, *Transparent predictions* (2013), https://papers.ssrn.com/sol3/papers.cfm?abstract_id=2324240 (last visited Mar 4, 2017); Latanya Sweeney, *Discrimination in online ad delivery*, 11 QUEUE 10 (2013); Frederik Zuiderveen Borgesius, *Algorithmic Decision-Making, Price Discrimination, and European Non-discrimination Law*, EUR. BUS. LAW REV. FORTHCOM. (2019); FRANK PASQUALE, THE BLACK BOX SOCIETY: THE SECRET ALGORITHMS THAT CONTROL MONEY AND INFORMATION (2015); VIKTOR MAYER-SCHÖNBERGER, DELETE: THE VIRTUE OF FORGETTING IN THE DIGITAL AGE (2011); Jeremias Prassl & Martin Risak, *Uber, taskrabbit, and co.: Platforms as employers-rethinking the legal analysis of crowdwork*, 37 COMP LAB POL J 619 (2015); O'NEIL, *supra* note; Kate Crawford & Ryan Calo, *There is a blind spot in AI research*, 538 NATURE 311–313 (2016); S. C. Olhede & P. J. Wolfe, *The growing ubiquity of algorithms in society: implications, impacts and innovations*, 376 PHIL TRANS R SOC A 20170364 (2018); Scott R. Peppet, *Regulating the internet of things: first steps toward managing discrimination, privacy, security and consent*, 93 TEX REV 85–176 (2014); Paul Ohm & Scott Peppet, *What If Everything Reveals Everything?*, BIG DATA MONOLITH MIT PRESS 2016 (2016); Omer Tene & Jules Polonetsky, *Taming the Golem: Challenges of ethical algorithmic decision-making*, 19 NCJL TECH 125 (2017).

[5] Among others Solon Barocas & Andrew D. Selbst, *Big data's disparate impact*, 104 CALIF. LAW REV. (2016); Pauline T. Kim, *Data-driven discrimination at work*, 58 WM MARY REV 857 (2016); Crystal Yang & Will Dobbie, *Equal Protection Under Algorithms: A New Statistical and Legal Framework*, AVAILABLE SSRN 3462379 (2019); Zach Harned & Hanna Wallach, *Stretching Human Laws to Apply to Machines: The Dangers of a 'Colorblind' Computer*, FLA. STATE UNIV. LAW REV. FORTHCOM. (2019); Thomas Nachbar, *Algorithmic Fairness, Algorithmic Discrimination*, VA. PUBLIC LAW LEG. THEORY RES. PAP. (2020).

In part, this legal ambiguity is unsurprising. The relevant legislation is written at a very high level of generality to allow for agile application across Member States. This intentional agility has left open the question of which tests should be used to assess discrimination in practice. Hard requirements for evidence of harm or strict statistical thresholds for legally acceptable disparity are avoided as the magnitude of discrimination depends largely on the type of harm committed, whether a minority group is affected, and whether the discrimination reflects a systemic injustice measured against the legal and political background of the relevant Member State(s). The admissibility and relevance of statistical tests, the make-up of disadvantaged and comparator groups, and the potential justifications for indirect discrimination and disparity across groups are traditionally decided on a case-by-case basis;[6] we refer to this approach as 'contextual equality'. Translating the law's high-level requirements into practical tests and measures is left to national and European courts that can account for such contextual considerations. While allowing for agility, this approach produces contradictory metrics and statistical tests (in the rare cases they are actually used) and a fragmented standard of protection across Europe.[7]

This heterogeneity in the interpretation and application of EU non-discrimination law, whilst desirable, poses a problem for building considerations of fairness and discrimination into automated systems. While numerous statistical metrics exist in the technical literature,[8] none can yet reliably capture a European conceptualisation of discrimination which is, by definition, contextual. Scalable automated methods to detect and combat discriminatory decision-making seemingly require clear-cut rules or quantifiable thresholds which

---

[6] Christopher McCrudden & Sacha Prechal, *The Concepts of Equality and Non-discrimination in Europe: A practical approach*, 2 EUR. COMM. DIR.-GEN. EMPLOY. SOC. AFF. EQUAL OPPOR. UNIT G (2009).

[7] LILLA FARKAS & DECLAN O'DEMPSEY, *How to present a discrimination claim: handbook on seeking remedies under the EU non-discrimination directives* 37 (2011); LILLA FARKAS ET AL., *Reversing the burden of proof: practical dilemmas at the European and national level* 37 (2015), http://dx.publications.europa.eu/10.2838/05358 (last visited Feb 9, 2020).

[8] Sahil Verma & Julia Rubin, *Fairness definitions explained, in* 2018 IEEE/ACM INTERNATIONAL WORKSHOP ON SOFTWARE FAIRNESS (FAIRWARE) 1–7 (2018); Sam Corbett-Davies & Sharad Goel, *The measure and mismeasure of fairness: A critical review of fair machine learning*, ARXIV PREPR. ARXIV180800023 (2018); Sorelle A. Friedler et al., *A comparative study of fairness-enhancing interventions in machine learning, in* PROCEEDINGS OF THE CONFERENCE ON FAIRNESS, ACCOUNTABILITY, AND TRANSPARENCY 329–338 (2019); Matt J Kusner et al., *Counterfactual Fairness, in* ADVANCES IN NEURAL INFORMATION PROCESSING SYSTEMS 30 4066–4076 (I. Guyon et al. eds., 2017), http://papers.nips.cc/paper/6995-counterfactual-fairness.pdf (last visited Jul 17, 2019); Geoff Pleiss et al., *On fairness and calibration, in* ADVANCES IN NEURAL INFORMATION PROCESSING SYSTEMS 5680–5689 (2017).

European non-discrimination law and jurisprudence purposefully do not provide. Judicial interpretive flexibility is not a 'bug' of EU non-discrimination law; rather, it is both intentional and advantageous. Nonetheless, in this regulatory environment, contextual agility can be quickly become runaway subjectivity, drastically tipping the balance of power in favour of would-be discriminators free to design systems and tests according to fairness metrics most favourable to them. Contextual equality does not lend itself to automation.

Even if standard metrics and thresholds were to emerge in European jurisprudence, problems remain. Cases have historically been brought against actions and policies that are potentially discriminatory in an intuitive or obvious sense. Compared to human decision-making, algorithms are not similarly intuitive; they operate at speeds, scale and levels of complexity that defy human understanding,[9] group and act upon classes of people that need not resemble historically protected groups,[10] and do so without potential victims ever being aware of the scope and effects of automated decision-making. As a result, individuals may never be aware they have been disadvantaged and thus lack a starting point to raise a claim under non-discrimination law.

These characteristics mean that intuition can no longer be relied upon, as it has been historically, as the primary mechanism to identify and assess potentially discriminatory actions in society. Algorithmic systems render a fundamental mechanism of EU non-discrimination law useless, necessitating new detection methods and evidential requirements. To fill this gap, we propose summary statistics that describe 'conditional demographic disparity' (CDD) as a static, baseline fairness metric that is harmonious with the 'gold standard' set by the Court of Justice for assessing potential discrimination. Based on our analysis, we argue that CDD can be used as a baseline measure to detect possible discrimination in automated systems that is both philosophically sound and harmonious with EU non-discrimination law and jurisprudence. Thus, we clarify why fairness cannot and should not be automated, and propose CDD as a baseline for evidence to ensure a consistent procedure for assessment (but not interpretation) across cases involving potential discrimination caused by automated systems.

---

This paper makes three contributions. First, we review the evidential requirements to bring a claim under EU non-discrimination law. Due to the disparate nature of algorithmic and human discrimination, the EU's current requirements are too contextual and open to judicial interpretation to be automated. Many of the concepts fundamental to bringing a claim, such as the composition of the disadvantaged and advantaged group or the severity and type of harm suffered, require normative or political choices to be made by the judiciary on a case-by-case basis.

Second, we show that automating fairness or non-discrimination in Europe may be impossible because the law does not provide a static or homogenous framework suited to testing for discrimination in AI systems. AI does not discriminate in an equivalent way to humans which disrupts established methods for detecting, investigating, and preventing discrimination. To contend with automated discrimination, we encourage the judiciary and industry to move away from predominantly using measures based on intuition, and towards a more coherent and consistent set of assessment procedures (not consistent interpretation) for automated discrimination. A 'gold standard' for assessment of *prima facie* discrimination has been advanced by the European Court of Justice but not yet translated into standard assessment procedures for automated discrimination.

Finally, we examine how existing work on fairness in machine learning lines up with procedures for assessing cases under EU non-discrimination law. We propose CDD as a standard baseline statistical measure that aligns with the ECJ's 'gold standard'. Establishing a standard set of statistical evidence for automated discrimination cases can help ensure consistent procedures for assessment, but not judicial interpretation, of cases involving AI and automated systems. Through this proposal for procedural regularity in the identification and assessment of automated discrimination, we clarify how to build considerations of fairness into automated systems as far as possible while still respecting and enabling the contextual approach to judicial interpretation practiced under EU non-discrimination law. Adoption of CDD will help ensure discriminatory thresholds and fairness metrics are not arbitrarily chosen and 'frozen' in code,[11] which would unjustifiably subvert case-specific judicial interpretation of non-discrimination law and implicitly shift this power to system developers.[12]

---

[11] LAWRENCE LESSIG, CODE: AND OTHER LAWS OF CYBERSPACE (2009).

[12] Dan L. Burk, *Algorithmic Fair Use*, 86 U CHI REV 283 (2019) addresses related problems in relation to legal rules and standards concerning copyright and algorithms; for an

Our aim is to increase dialogue between the legal and technology communities in order to create legally sound and scalable solutions for fairness and non-discrimination in automated systems in Europe. We use 'automated fairness' as shorthand to refer to a plethora interdisciplinary work meant to embed considerations of fairness into the design and governance of automated systems. Major categories of this foundational work include the development of statistical metrics for fairness[13] and their societal implications,[14] bias testing using sensitive data,[15] prevention of bias via causal reasoning,[16] testing for disparate impact,[17] due process rules for algorithms,[18] software toolkits to analyse models and datasets for bias,[19] developer and institutional codes of conduct, checklists, and impact assessment forms[20] as well as and assessment of their utility to practitioners,[21] standardised documentation for models and training datasets,[22] and related mechanisms to take the 'human out of the loop' and eliminate human discretion and subjectivity in automated decision-making. Along with these technical, empirical, and organisational tools and measures, complementary duties have been proposed for system controllers and developers including a duty of care for online harms[23] and fiduciary duties for technology companies.[24] Much of this critical work is organised within the FAT* (Fairness, Accountability, and Transparency in Machine Learning) research network.

To address the challenges of automated discrimination, it is important that such communities and the legal community work together and learn from each other. We call on the legal community to draw inspiration from technologists and their coherent approaches and consistency. At the same time, we encourage technologists to embrace flexibility and acknowledge the idea of contextual equality. CDD used as a statistical baseline measure will be of service to judges, regulators, industry or claimants. Judges will have a first frame to investigate *prima facie* discrimination, regulators will have a baseline for investigating potential discrimination cases, industry can prevent discrimination or refute potential claims, and victims will have a reliable measure of potential discrimination to raise claims.

---

*global landscape of AI ethics guidelines*, 1 NAT. MACH. INTELL. 389–399 (2019); Jessica Fjeld et al., *Principled Artificial Intelligence: Mapping Consensus in Ethical and Rights-based Approaches to Principles for AI*, BERKMAN KLEIN CENT. RES. PUBL. (2020); Urs Gasser & Carolyn Schmitt, *The Role of Professional Norms in the Governance of Artificial Intelligence* (2019); Brent Mittelstadt, *Principles alone cannot guarantee ethical AI*, 1 NAT. MACH. INTELL. 501–507 (2019).

[21] Kenneth Holstein et al., *Improving fairness in machine learning systems: What do industry practitioners need?*, in PROCEEDINGS OF THE 2019 CHI CONFERENCE ON HUMAN FACTORS IN COMPUTING SYSTEMS 1–16 (2019).

[22] Timnit Gebru et al., *Datasheets for Datasets* (2018), https://arxiv.org/abs/1803.09010 (last visited Oct 1, 2018); Margaret Mitchell et al., *Model Cards for Model Reporting*, PROC. CONF. FAIRNESS ACCOUNT. TRANSPAR. - FAT 19 220–229 (2019); Sarah Holland et al., *The Dataset Nutrition Label: A Framework To Drive Higher Data Quality Standards*, ARXIV180503677 CS (2018), http://arxiv.org/abs/1805.03677 (last visited Oct 1, 2018).

[23] Woods Lorna & Perrin William, *An updated proposal by Professor Lorna Woods and William Perrin*, https://d1ssu070pg2v9i.cloudfront.net/pex/carnegie_uk_trust/2019/01/29121025/Internet-Harm-Reduction-final.pdf (last visited May 11, 2019).

[24] Technology | Academics | Policy - Jonathan Zittrain and Jack Balkin Propose Information Fiduciaries to Protect Individual Privacy Rights, , http://www.techpolicy.com/Blog/September-2018/Jonathan-Zittrain-and-Jack-Balkin-Propose-Informat.aspx (last visited Feb 2, 2019); Jack M. Balkin, *Information Fiduciaries and the First Amendment Lecture*, 49 UC DAVIS LAW REV. 1183–1234 (2015).

## II.   THE UNIQUE CHALLENGE OF AUTOMATED DISCRIMINATION

Artificial intelligence creates new challenges for establishing *prima facie* discrimination. By definition claimants must experience or anticipate inequality. Compared to traditional forms of discrimination, automated discrimination is more abstract and unintuitive, subtle, and intangible.[25] These characteristics make it difficult to detect and prove as victims may never realise they have been disadvantaged.[26] Seeing colleagues getting hired or promoted, or comparing prices in supermarkets, help us to understand whether we are treated fairly. In an algorithmic world this comparative element is increasingly eroded; it will be much harder, for example, for consumers to assess whether they have been offered the best price possible or to know that certain advertisements have not been shown to them.[27]

Intuitive or *prima facie* experiences of discrimination essential to bringing claims under EU non-discrimination law are diminished. Although experiences of discrimination are likely to diminish, the same cannot be said of discriminatory practices. This is a problem, as knowing where to look and obtaining relevant evidence that could reveal *prima facie* discrimination will be difficult when automated discrimination is not directly experienced or 'felt' by potential claimants, and when access to (information about) the system is limited.[28] System controllers may, for example, limit the availability of relevant evidence to protect their intellectual property or avoid litigation.[29] Caution must be exercised as explicit (e.g. not to promote women) and implicit bias (e.g. only to promote people who play football) are recorded in the data used to train and run AI systems[30] but lack the intuitive feeling of inequality.

---

[25] Brent Mittelstadt et al., *The ethics of algorithms: Mapping the debate*, 3 BIG DATA SOC. (2016), http://bds.sagepub.com/lookup/doi/10.1177/2053951716679679 (last visited Dec 15, 2016); Wachter, *supra* note 10 at 42–43, 45–46; Tal Zarsky, *The Trouble with Algorithmic Decisions An Analytic Road Map to Examine Efficiency and Fairness in Automated and Opaque Decision Making*, 41 SCI. TECHNOL. HUM. VALUES 118–132 (2016).

[26] Wachter, *supra* note 10.

[27] *Id.*

[28] Brent Mittelstadt, *Automation, Algorithms, and Politics| Auditing for Transparency in Content Personalization Systems*, 10 INT. J. COMMUN. 12 (2016).

[29] Burrell, *supra* note 9; Sandra Wachter, Brent Mittelstadt & Chris Russell, *Counterfactual Explanations without Opening the Black Box: Automated Decisions and the GDPR*, 3 HARV. J. LAW TECHNOL. 841–887 (2018); Sandra Wachter, Brent Mittelstadt & Luciano Floridi, *Why a Right to Explanation of Automated Decision-Making Does Not Exist in the General Data Protection Regulation*, 7 INT. DATA PRIV. LAW 76–99 (2017); Jeremy B. Merrill Ariana Tobin, *Facebook Is Letting Job Advertisers Target Only Men*, PROPUBLICA (2018), https://www.propublica.org/article/facebook-is-letting-job-advertisers-target-only-men (last visited Mar 24, 2019); ProPublica Data Store, *COMPAS Recidivism Risk Score Data and Analysis*, PROPUBLICA DATA STORE (2016), https://www.propublica.org/datastore/dataset/compas-recidivism-risk-score-data-and-analysis (last visited May 7, 2019).

Similarly, intuition might fail us when evaluation bias because the data and procedures used do make our alarm bells ring. It cannot be assumed that automated systems will discriminate in ways similar to humans, or familiar known patterns of discrimination.[31] AI and automated systems are valued precisely because of their ability to process data at scale and find unintuitive connections and patterns between people.[32] These systems will stratify populations and outcomes according to the data and features they are fed, classification rules they are given and create, and implicit biases in the data and system design.[33]

On the one hand, AI poses a challenge to protecting legally protected groups: new and counterintuitive proxies for traditionally protected characteristics will emerge but not necessarily be detected.[34] On the other hand, AI poses a challenge to the scope of non-discrimination law itself. It cannot be assumed that disparity will occur only between legally protected groups.

Groups which do not map to a legally protected characteristics may suffer levels of disparity which would otherwise be considered discriminatory if applied to a protected group.[35] These new patterns of disparity may force legislators and society to re-consider whether the scope of non-discrimination remains broad enough to capture significant disparity as caused not only by humans and organisations, but machines as well.[36]

---

[30] On how biased data leads to biased outcomes see Alexandra Chouldechova, *Fair Prediction with Disparate Impact: A Study of Bias in Recidivism Prediction Instruments*, 5 BIG DATA 153–163 (2017); see also Jerry Kang et al., *Implicit bias in the courtroom*, 59 UCLA REV 1124 (2011); Marion Oswald & Alexander Babuta, *Data Analytics and Algorithmic Bias in Policing* (2019).

[31] Sandra Wachter & B. D. Mittelstadt, *A right to reasonable inferences: re-thinking data protection law in the age of Big Data and AI*, 2019 COLUMBIA BUS. LAW REV. (2019); Timo Makkonen, *Equal in law, unequal in fact: racial and ethnic discrimination and the legal response thereto in Europe*, 2010; Jennifer Cobbe & Jatinder Singh, *Regulating Recommending: Motivations, Considerations, and Principles*, forthcoming EUR. J. LAW TECHNOL. (2019), https://papers.ssrn.com/abstract=3371830 (last visited Feb 28, 2020).

[32] Luciano Floridi, *The search for small patterns in big data*, 2012 PHILOS. MAG. 17–18 (2012).

[33] Batya Friedman & Helen Nissenbaum, *Bias in computer systems*, 14 ACM TRANS. INF. SYST. TOIS 330–347 (1996); Kate Crawford, *The hidden biases in big data*, 1 HBR BLOG NETW. (2013).

[34] Anupam Datta et al., *Proxy Non-Discrimination in Data-Driven Systems*, ARXIV PREPR. ARXIV170708120 (2017), https://arxiv.org/abs/1707.08120 (last visited Sep 22, 2017); Barocas and Selbst, *supra* note 5; Brent Mittelstadt & Luciano Floridi, *The Ethics of Big Data: Current and Foreseeable Issues in Biomedical Contexts*, 22 SCI. ENG. ETHICS 303–341 (2016).

[35] Mittelstadt, *supra* note 10; 126 LINNET TAYLOR, LUCIANO FLORIDI & BART VAN DER SLOOT, GROUP PRIVACY: NEW CHALLENGES OF DATA TECHNOLOGIES (2016); Alessandro Mantelero, *From Group Privacy to Collective Privacy: Towards a New Dimension of Privacy and Data Protection in the Big Data Era*, in GROUP PRIVACY 139–158 (2017); LEE A. BYGRAVE, DATA PROTECTION LAW: APPROACHING ITS RATIONALE, LOGIC AND LIMITS (2002).

Another fundamental challenge arises from the non-decomposable nature of complex modern machine learning or AI systems. While indirect discrimination has been detected in complex systems such as, for example, governmental provision housing in Hungary,[37] this has in no small part has been possible because of the ability of the judiciary to decompose these large systems into isolated components or a collection of housing policies, and to evaluate whether each such policy satisfies non-discrimination law. This decomposition has been an incredibly effective tool that has allowed judges to bring 'common sense reasoning' to bare on complex and nuanced systems, and also strengthens the power of statistical reasoning, making the choice of test less significant (see: discussion of test choice in Sections V and VI). However, attempting to understand discrimination caused by complex algorithmic systems that are not similarly decomposable reveals the limitations of common-sense reasoning.

These limitations suggest that, at a minimum, automated discrimination will be difficult to prove without readily available and relevant statistical evidence.[38] Claimants may lack information regarding a system's optimisation conditions or decision rules, and thus be unaware of the reach and definition of the contested rule that led to perceived disparity.[39] Similarly, without information regarding the scope of the system and the outputs or decisions received

---

[36] Wachter and Mittelstadt, *supra* note 31.

[37] Equal Treatment Authority, Case EBH/67/22/2015 (2015), http://egyenlobanas-mod.hu/article/view/ebh-67-2015.

[38] TIMO MAKKONEN, MEASURING DISCRIMINATION DATA COLLECTION AND EU EQUALITY LAW 28 (2007).

[39] Wachter, *supra* note 10 at 45–46 provides an example of this difficulty compared with 'analogue' discrimination based on a case heard by the Federal Labour Court in Germany.: "If for example an employer decides that only people taller than 1 meter 80 cm should be hired, it will be easy to establish prima facie discrimination. In this case we know the rule (hiring strategy) and can find statistical evidence to show that, while the rule does not directly use gender as a discriminating factor, it would nonetheless affect women disproportionately. In the online world, we often do not know the rules and attributes on which we are profiled and whether these attributes correlate with protected characteristics. Further, we do not know who else is in our profiling group, which other groups exist, and how we are treated in comparison to others. This makes it difficult to prove that a protected group was disproportionally negatively affected." The example is inspired by case 8 AZR 638/14 heard by the Federal Labour Court in Germany on 18 February 2016 which addressed Lufthansa's policy of only hiring airline pilots taller than 1.65 meters.

by other individuals,[40] claimants will have difficulty defining a legitimate comparator group.

## III.      CONTEXTUAL EQUALITY IN EU NON-DISCRIMINATION LAW

European non-discrimination law exists in both primary[41] and secondary law.[42] The widest scope of non-discrimination law exists in primary law in Article 21 of the EU Charter of Fundamental Rights. Article 21 establishes that "[a]ny discrimination based on any ground such as sex, race, colour, ethnic or social origin, genetic features, language, religion or belief, political or any other opinion, membership of a national minority, property, birth, disability, age or sexual orientation shall be prohibited." This is a non-exclusive list that is sector-neutral. However, the Charter only applies to public bodies of the European Union and the Member States, and not the private sector.[43]

In contrast, secondary law applies to both the private and the public sectors. In this paper we will focus on the four non-discrimination directives of the EU:[44] the Racial Equality Directive (2000/43/EC),[45] the Gender Equality

---

[40] Christian Sandvig et al., *Auditing algorithms: Research methods for detecting discrimination on internet platforms*, DATA DISCRIM. CONVERT. CRIT. CONCERNS PRODUCT. INQ. (2014), http://social.cs.uiuc.edu/papers/pdfs/ICA2014-Sandvig.pdf (last visited Feb 13, 2016); Mittelstadt, *supra* note 28.

[41] EVELYN ELLIS & PHILIPPA WATSON, EU ANTI-DISCRIMINATION LAW at 13 (2012); for a fantastic overview of the scope, history, and effectiveness of EU non-discrimination law see SANDRA FREDMAN, DISCRIMINATION LAW (2011); and also MARK BELL, ANTI-DISCRIMINATION LAW AND THE EUROPEAN UNION (2002).

[42] ELLIS AND WATSON, *supra* note 41 at at 19-21.

[43] It should be noted that in the cases *Mangold* and *Egenberger* the ECJ opened up the possibility of Article 21 being applicable between private actors, albeit in very rare, exceptional and limited circumstances. To date the ECJ has not heard a case where this was applicable.

[44] European Commission, *Non-discrimination*, EUROPEAN COMMISSION - EUROPEAN COMMISSION (2020), https://ec.europa.eu/info/aid-development-cooperation-fundamental-rights/your-rights-eu/know-your-rights/equality/non-discrimination_en (last visited Mar 3, 2020).

[45] COUNCIL DIRECTIVE 2000/43/EC OF 29 JUNE 2000 IMPLEMENTING THE PRINCIPLE OF EQUAL TREATMENT BETWEEN PERSONS IRRESPECTIVE OF RACIAL OR ETHNIC ORIGIN, , OJ L 180 (2000), http://data.europa.eu/eli/dir/2000/43/oj/eng (last visited Aug 5, 2019).

Directive (recast) (2006/54/EC),[46] the Gender Access Directive (2004/113/EC),[47] and the Employment Directive (2000/78/EC).[48]

The scope of groups and sectors protected varies across these four directives. Article 3 of the Racial Equality Directive prohibits discrimination based on race or ethnicity in the context of employment, access to the welfare system, social protection, education, as well as goods and services. The Gender Equality Directive establishes equality in the workplace, but equal treatment is only guaranteed for social security and not in the broader welfare system, including social protection and access to healthcare and education.[49] Similarly the Gender Access Directive provides equal access to goods and services but excludes media content, advertisements, and education from its scope.[50] The Employment Directive only prevents discrimination based on religion or beliefs, disability, age, and sexual orientation in the workplace, but does not guarantee access to goods and services or the welfare system.[51]

As is the case with all EU directives as (opposed to EU regulations), the non-discrimination directives only establish a minimal standard and provide a general framework that needs to be transposed into national law by the

---

[46] DIRECTIVE 2006/54/EC OF THE EUROPEAN PARLIAMENT AND OF THE COUNCIL OF 5 JULY 2006 ON THE IMPLEMENTATION OF THE PRINCIPLE OF EQUAL OPPORTUNITIES AND EQUAL TREATMENT OF MEN AND WOMEN IN MATTERS OF EMPLOYMENT AND OCCUPATION (RECAST), , OJ L 204 (2006), http://data.europa.eu/eli/dir/2006/54/oj/eng (last visited Aug 5, 2019).

[47] COUNCIL DIRECTIVE 2004/113/EC OF 13 DECEMBER 2004 IMPLEMENTING THE PRINCIPLE OF EQUAL TREATMENT BETWEEN MEN AND WOMEN IN THE ACCESS TO AND SUPPLY OF GOODS AND SERVICES, , OJ L 373 (2004), http://data.europa.eu/eli/dir/2004/113/oj/eng (last visited Aug 5, 2019).

[48] COUNCIL DIRECTIVE 2000/78/EC OF 27 NOVEMBER 2000 ESTABLISHING A GENERAL FRAMEWORK FOR EQUAL TREATMENT IN EMPLOYMENT AND OCCUPATION, , OJ L 303 (2000), http://data.europa.eu/eli/dir/2000/78/oj/eng (last visited Aug 5, 2019).

[49] EUROPEAN UNION AGENCY FOR FUNDAMENTAL RIGHTS AND COUNCIL OF EUROPE, HANDBOOK ON EUROPEAN NON-DISCRIMINATION LAW at 22 (2018 edition ed. 2018), https://fra.europa.eu/sites/default/files/fra_uploads/1510-fra-case-law-handbook_en.pdf.

[50] Other legitimate aims that limit the scope of the Directive are named in Recital 16: "[d]ifferences in treatment may be accepted only if they are justified by a legitimate aim. A legitimate aim may, for example, be the protection of victims of sex-related violence (in cases such as the establishment of single sex shelters), reasons of privacy and decency (in cases such as the provision of accommodation by a person in a part of that person's home), the promotion of gender equality or of the interests of men or women (for example single-sex voluntary bodies), the freedom of association (in cases of membership of single-sex private clubs), and the organisation of sporting activities (for example single-sex sports events)."

[51] For further discussion on the limited scope of these directives and its implications see Wachter, *supra* note 10.

Member States. Many Member States offer higher protection then the minimal standards set out by the directives.[52] This variability leaves Europe with a fragmented standard across the Member States. [53]

European non-discrimination law addresses two general types of discrimination: direct and indirect.[54] Direct discrimination refers to adverse treatment based on a protected attribute such as sexual orientation or gender. Indirect discrimination on the other hand describes a situation where "apparently neutral provision, criterion or practice"[55] disproportionately disadvantages a protected group in comparison with other people.

One advantage of indirect discrimination over direct discrimination is that, whereas the former focuses primary on individual cases of discrimination, the latter deals with rules or patterns of behaviour and can thus reveal underlying social inequalities. Indirect discrimination can thus help to shed light on systematic and structural unfairness in a society and advocate for social change.[56] With that said, indirect discrimination is a relatively new concept in European Member States meaning there is a relative lack of case law addressing the concept.[57]

To bring a case alleging direct or indirect discrimination under EU non-discrimination law a claimant must meet several evidential requirements that together establish *prima facie* discrimination. Claimants must demonstrate that (1) a particular harm has occurred or is likely to occur; (2) the harm manifests or is likely to manifest significantly within a protected group of people; and (3) the harm is disproportionate when compared with others in a similar situation. Once these requirements are met the burden of proof shifts to the alleged

---

[52] See EUROPEAN PARLIAMENT'S DIRECTORATE-GENERAL FOR PARLIAMENTARY RESEARCH SERVICES, *Gender Equal Access to Goods and Services Directive 2004/113/EC -European Implementation Assessment* at I-38, http://www.europarl.europa.eu/RegData/etudes/STUD/2017/593787/EPRS_STU(2017)593787_EN.pdf (last visited Mar 26, 2019) for an overview of how the Member States have implemented the framework.

[53] For an overview of the fragmented standards across the EU Member States see EUROPEAN NETWORK OF LEGAL EXPERTS IN GENDER EQUALITY AND NON-DISCRIMINATION, *A comparative analysis of non-discrimination law in Europe* (2018), doi:10.2838/939337.

[54] ELLIS AND WATSON, *supra* note 41 at 142.

[55] This is stated in all EU Non-Discrimination Directives, see also Christopher McCrudden, *The New Architecture of EU Equality Law after CHEZ: Did the Court of Justice Reconceptualise Direct and Indirect Discrimination?*, EUR. EQUAL. LAW REV. FORTHCOM., at 3 (2016).

[56] McCrudden and Prechal, *supra* note 6 at 35.

[57] MAKKONEN, *supra* note 38 at 33. The exception to this trend is the United Kingdom, where the concept has a much longer history and features in relatively more case law compared to other Member States.

offender who can then justify the contested rule or practice, or otherwise refute the claim.[58]

The following sections examine how these evidential requirements have been interpreted in jurisprudence of the European Court of Justice and national courts of the Member States. The jurisprudence reveals that defining a disadvantaged group(s), legitimate comparator group(s), and evidence of a 'particular disadvantage' requires the judiciary to make case-specific normative choices that reflect local political, social, and legal dimensions of the case as well as arguments made by claimants and alleged offenders.[59] The reviewed jurisprudence reveals very few clear-cut examples of static rules, requirements, or thresholds for defining the key concepts and groups underlying discrimination as a legal standard.

These normative decisions are not made in isolation or sequentially, but rather are interconnected and based in the facts of the case. This contextual normative flexibility, or 'contextual equality', is not a 'bug' or unintended application of non-discrimination law; rather, it is intentional and, barring significant regulatory and judicial re-alignment, must be respected and facilitated in automated systems.[60] This will not be a simple task. System developers and controllers have very little consistent guidance to draw on in designing considerations of fairness, bias, and non-discrimination into AI and automated systems. Replicating the judiciary's approach to 'contextual equality' will be difficult, if not impossible, to replicate in automated systems at scale.

A.    COMPOSITION OF THE DISADVANTAGED GROUP

'Discrimination' refers to an adverse act committed against a legally protected individual or group. A first step in bringing a claim is to define the group that has been disadvantaged. In direct discrimination this is a simple task: the rule, practice, or action alleged to be discriminatory must explicitly refer to a protected characteristic. For indirect discrimination, defining the disadvantaged group is more complicated: an "apparently neutral provision, criterion or practice" must be shown to significantly disadvantage a legally protected group despite not explicitly addressing this group. Direct discrimination can thus be proven at an individual level through explicit reference to a protected characteristic in the contested rule, whereas indirect discrimination requires group-level comparison.[61]

---

[58] FARKAS ET AL., *supra* note 7 at 9.

[59] MAKKONEN, *supra* note 38 at 36.

[60] This flexibility stems from the fact that the majority of EU level non-discrimination law is enacted through directives which require individual enactment, interpretation, and enforcement by Member states. This choice is unsurprising given the contextual nature of equality and fairness which are defined against cultural norms and expectations. For a discussion of the role of culture, history, and context in notions of equality, see: DOUGLAS W. RAE ET AL., EQUALITIES (1981).

To successfully bring an indirect discrimination case, the claimant must provide evidence that an "apparently neutral provision, criterion or practice" significantly disadvantages a protected group when compared with other people in a similar situation. The contested rule must be shown to have actually harmed a protected group or have the potential for significant harm. This raises the question of how the disadvantaged group is defined. In other words, what are the common characteristics of the group, and are these legally protected?[62]

Prior national and ECJ jurisprudence does not provide comprehensive rules for the composition of the disadvantaged group; rather, the composition of the disadvantaged group is set according to the facts of the case. As a rule of thumb, disadvantaged group(s) can be defined by 'broad' traits such as sexual orientation or religious beliefs, or by 'narrow' traits that describe specific demographic subgroups such as a specific subgroup of men[63] or groups such as blind persons, black people, or people younger than 40.[64] The appropriate level of abstraction or degree of difference between disadvantaged and comparator groups will be determined according to the facts of the case, normally accounting for the reach of the contested rule as well as potential comparators (see: Section C).

With that said, some consistency exists. First, the ECJ has suggested that the disadvantaged and comparator groups should be defined in relation to the contested rule as implemented by an alleged offender. In *Allonby*[65] the ECJ ruled that the composition of the disadvantaged group and the comparator group, or the "category of persons who may be included in the comparison," is determined by the "apparently neutral provision, criterion or practice" (i.e. the 'contested rule') in question.[66] The alleged offender is defined as the "single source […] which is responsible for the inequality and which could restore equal treatment."[67] This could, for example, be an employer,[68] legislator,[69] or trade union representative.[70] The reach of the contested rule as implemented by the alleged offender determines who could potentially be affected, and thus the composition of the disadvantaged groups and comparator groups.

---

[61] MAKKONEN, *supra* note 38 at 33.

[62] *Id.* at 36.

[63] Case C-104/09, Pedro Manuel Roca Álvarez v Sesa Start España ETT SA, 2010 E.C.R. I-08661, 103, https://eur-lex.europa.eu/legal-content/EN/TXT/?uri=CELEX%3A62009CJ0104 in which the ECJ accepted as unlawful discrimination the unfavourable treatment of a sub-category of men.

[64] MAKKONEN, *supra* note 38 at 36.

[65] Case C-256/01, Debra Allonby v Accrington & Rossendale College, Education Lecturing Services, trading as Protocol Professional and Secretary of State for Education and Employment., 2004 E.C.R. I–00873, 46, http://curia.europa.eu/juris/liste.jsf?language=en&num=C-256/01.

[66] *Id.* at 73.

The 'reach' of a contested rule depends on the facts of the case and arguments presented by the claimant and alleged offender. Prior jurisprudence and legal commentators point towards several potential factors to determine the reach of the contested rule including legislation of a member state,[71] contractual agreement within a company,[72] regional law,[73] collective agreement within

a particular sector,[74] as well as hiring, firing and promotion practices or payment strategies.[75]

An ad hoc approach to determining the reach of the contested rule, and thus the composition of the disadvantaged group, has traditionally worked for cases of intuitive discrimination. However, it is unlikely to be effective for cases of discrimination involving online platforms and services, or multi-national companies, where the reach of the potential offender is not self-evident.[76] What, for example, is the reach of Google, Facebook, and Amazon when they advertise products or rank search results?

Arguably Google, Facebook or Amazon's potential reach could be seen as global.[77] Amazon, for example, was accused of using biased hiring algorithms.[78] One could make the argument that a job advertised by Amazon has global reach and thus global statistics should be the basis to assess whether discrimination has occurred when targeting people. Another viable argument might be to only look at the people that have actually applied for the job, if decisions made by algorithms used to filter applicants are contested. Alternatively, it could be argued that only 'qualified' people should be considered, further limiting the statistics to applicants who meet a minimal set of requirements for the job.[79] Similar discussions have already emerged in competition law where it remains similarly unclear as to how the market reach of technology companies and digital platforms should be defined.[80] As these examples reveal, the reach of the contested rule depends on case-specific contextual factors and argumentation which will be difficult to embed in automated systems at scale.

---

[74] CASE C-127/92, *supra* note 67; Joined cases C-399/92, C-409/92, C-425/92, C-34/93, C-50/93 and C-78/93, Stadt Lengerich v Angelika Helmig and Waltraud Schmidt v Deutsche Angestellten-Krankenkasse and Elke Herzog v Arbeiter-Samariter-Bund Landverband Hamburg eV and Dagmar Lange v Bundesknappschaft Bochum and Angelika Kussfeld v Firma Detlef Bogdol GmbH and Ursula Ludewig v Kreis Segeberg, 1994 E.C.R. I-05727, https://eur-lex.europa.eu/legal-content/EN/TXT/?uri=CELEX%3A61992CJ0399.

[75] MAKKONEN, *supra* note 38 at 34.

[76] One how online ads can have discriminatory outcomes see Ariana Tobin Julia Angwin, *Facebook (Still) Letting Housing Advertisers Exclude Users by Race*, PROPUBLICA (2017), https://www.propublica.org/article/facebook-advertising-discrimination-housing-race-sex-national-origin (last visited Mar 24, 2019); Terry Parris Jr Julia Angwin, *Facebook Lets Advertisers Exclude Users by Race*, PROPUBLICA (2016), https://www.propublica.org/article/facebook-lets-advertisers-exclude-users-by-race (last visited Mar 24, 2019); Till Speicher et al., *Potential for Discrimination in Online Targeted Advertising Till Speicher MPI-SWS MPI-SWS MPI-SWS*, 81 *in* PROCEEDINGS OF THE CONFERENCE ON FAIRNESS, ACCOUNTABILITY, AND TRANSPARENCY (FAT*) 1–15 (2018); Muhammad Ali et al., *Discrimination through optimization: How Facebook's ad delivery can lead to skewed outcomes*, ARXIV PREPR. ARXIV190402095 (2019).

[77] However, while the reach of Google (for example) as a platform is arguably global, in specific cases it may be possible to specify a narrower reach based on the preferences or intended audience defined by the client or advertiser. It is in this sense that the reach of a contested rule must be determined contextually according to the facts of the case.

[78] Reuters, *Amazon ditched AI recruiting tool that favored men for technical jobs*, THE GUARDIAN, October 10, 2018, https://www.theguardian.com/technology/2018/oct/10/amazon-hiring-ai-gender-bias-recruiting-engine (last visited Mar 2, 2020).

### 1. *Multi-dimensional discrimination*

A related issue that arises when considering the composition of the disadvantaged group is the problem of multi-dimensional discrimination. This occurs when people are discriminated against based on more than one protected characteristic. A distinction can be drawn between additive and intersectional discrimination. Additive discrimination refers to disparity based on two or more protected characteristics considered individually, for example, being treated separately as "black" and a "woman." Intersectional discrimination refers to a disadvantage based on two or more characteristics considered together, for example being a "black woman." [81]

Whilst the non-discrimination Directive mentions the phenomenon of intersectional discrimination in its Recital 14 in reference to the fact that "women are often the victims of multiple discrimination,"[82] the ECJ has a more restrictive view. One particular judgement is problematic for both additive and intersectional discrimination. The *Parris* case centred on a claimant and his civil partner with regards to survival pension.[83] Mr. Parris tried to bring a

---

discrimination claim based on both age and sexual orientation. However, the Court explained that a contested rule is "not capable of creating discrimination as a result of the combined effect of sexual orientation and age, where that rule does not constitute discrimination either on the ground of sexual orientation or on the ground of age taken in isolation."[84] In other words, the two protected characteristics must be considered separately in assessing whether Mr. Parris suffered a particular disadvantage. The Court also explained that no new protected group or category of discrimination can be created based on the facts of the case.[85]

The judgement has been subject to criticism owing to its detrimental effect on people affected by intersectional and additive discrimination.[86] Some Member States have since acknowledged this legal challenge,[87] but case law dealing with multi-dimensional discrimination remains scarce. National provisions have not yet yielded a successful case in this regard.[88] The lack of jurisprudence addressing discrimination based on multiple factors reflects the lack of clear guidance which can be applied to the development and governance of automated systems.

## B. COMPOSITION OF THE COMPARATOR GROUP

EU non-discrimination law is inherently comparative in both indirect and direct (albeit to a lesser degree) discrimination cases. Illegal disparity occurs when a protected group is treated less favourably than others in a similar situation. This definition follows from a basic principle of non-discrimination law stemming from an Aristotelian conception of equality: treat like cases alike and different cases differently.[89] This guiding principle creates a requirement to identify a 'legitimate comparator', or an individual or group which has been

---

60th birthday - despite being in a relationship of 30 years - because civil partnerships were not yet legal.

[84] *Id.* at 83(3).

[85] *Id.* at 80.

[86] Erica Howard, *EU anti-discrimination law: Has the CJEU stopped moving forward?*, 18 INT. J. DISCRIM. LAW 60–81 (2018); Raphaële Xenidis, *Multiple discrimination in EU anti-discrimination law: towards redressing complex inequality?* (2018).

[87] ISABELLE CHOPIN, CARMINE CONTE & EDITH CHAMBRIER, *A comparative analysis of non-discrimination law in Europe 2018* 46–47 (2018), https://www.equalitylaw.eu/downloads/4804-a-comparative-analysis-of-non-discrimination-law-in-europe-2018-pdf-1-02-mb. In Austria, for example, the law allows higher damages if discrimination on prohibited grounds occurred.

[88] *Id.* at 48.

[89] McCrudden and Prechal, *supra* note 6 at 11–13; ARISTOTLE, THE COMPLETE WORKS OF ARISTOTLE: THE REVISED OXFORD TRANSLATION, ONE-VOLUME DIGITAL EDITION Nicomachean Ethics 3834 (Jonathan Barnes ed., 2014).

unjustifiably treated better than an individual or group in a comparable situation. Identification of a legitimate comparator is essential to bring a successful case under EU non-discrimination law.[90]

EU non-discrimination directives do not clearly indicate requirements for the composition of the advantaged comparator group.[91] Nonetheless, a comparator must be identified as "the Court has consistently held that the principle of equal treatment requires that comparable situations must not be treated differently, and different situations must not be treated in the same way, unless such treatment is objectively justified."[92] A comparison between groups is implicit and required by the principle of equal treatment.[93]

In identifying comparator groups, two practical requirements stand out. First, who can a claimant legitimately compare themselves with, and what argumentation or evidence is required to establish that the two groups should be seen as equal and in a similar situation? Second, must a concrete comparator be identified to establish *prima facie* discrimination, or is a hypothetical comparator sufficient? The agility and context-specificity of these requirements pose a challenge for automating fairness.

As with disadvantaged groups, comparators are case-specific and defined against the questionable practice or rule being challenged (the 'facts of the case'). Prior cases heard by the ECJ have addressed questions such as:

- Are married couples equal to same-sex civil partnerships?[94]
- Are clinical therapists and pharmacists performing equal tasks?[95]
- Is leaving a job after three years for important reasons (not including pregnancy) the same as leaving a job after five years for "other reasons" including pregnancy?[96]

---

[90] McCrudden and Prechal, *supra* note 6 at 13.

[91] MAKKONEN, *supra* note 38 at 36.

[92] Case C-127/07, Société Arcelor Atlantique et Lorraine and Others v Premier ministre, Ministre de l'Écologie et du Développement durable and Ministre de l'Économie, des Finances et de l'Industrie, 2008 E.C.R. I-09895, 23, http://curia.europa.eu/juris/liste.jsf?language=en&num=C-127/07.

[93] McCrudden and Prechal, *supra* note 6 at 12; the principles of equal treatment is established and defined in Articles 1(1) and 3(1) of the EUROPEAN COUNCIL, *Council Directive 76/207/EEC of 9 February 1976 on the implementation of the principle of equal treatment for men and women as regards access to employment, vocational training and promotion, and working conditions* 76 (1976), https://eur-lex.europa.eu/legal-content/EN/TXT/?uri=CELEX%3A31976L0207.

[94] ELLIS AND WATSON, *supra* note 81 at 102.

[95] CASE C-127/92, *supra* note 67.

[96] CASE C-1/95, *supra* note 73.

- Are full-time workers comparable to part-time workers?[97]
- Is an education from a 'third country' comparable with one from an EU Member State?[98]

Answering such questions is not straightforward. Simple binary comparisons are rare. A key challenge for claimants is that many attributes traditionally considered sensitive such as gender, ethnicity, or disability are "social constructs, which entails that their understanding is complex and dependent on a given social context." Legitimate comparators for such constructs are necessarily also bound by social context.[99]

Further complicating matters, socially constructed characteristics are not consistently protected at a European level[100] because Directives must be interpreted and transposed into national law. Under the EU gender equality directives[101] gender discrimination is prohibited based on biological sex and according to the ECJ also based on transsexuality (i.e. gender reassignment).[102] However, the standard of the protection of the broader concept of 'gender identity' remains an open question and has been interpreted differently across the Member States resulting in a fragmented standard.[103] Other grounds such as ethnicity in the Racial Equality Directive, and disability, religion, and sexual orientation in the Employment Directive are also interpreted, defined and protected through Member state law, where their scope and definition varies greatly. For example, some Member State laws protect Scientology as a religion, whereas others do not. Similarly, national laws vary regarding whether transgender should be considered a type of 'sex' or 'sexual orientation'.[104]

---

[97] JOINED CASES C-399/92, C-409/92, C-425/92, C-34/93, C-50/93 AND C-78/93, *supra* note 74.

[98] Case C-457/17, Heiko Jonny Maniero v Studienstiftung des deutschen Volkes eV, 2018 ECLI:EU:C:2018:912, https://eur-lex.europa.eu/legal-content/EN/TXT/?uri=CELEX%3A62017CN0457; Case C-703/17 Opinion of Advocate General Leger, Adelheid Krah v Universität Wien, 2019 ECLI:EU:C:2019:450, https://eur-lex.europa.eu/legal-content/EN/TXT/?uri=CELEX:62017CC0703 (last visited Mar 26, 2019).

[99] FARKAS ET AL., *supra* note 7 at 41.

[100] *Id.* at 42.

[101] DIRECTIVE 2006/54/EC OF THE EUROPEAN PARLIAMENT AND OF THE COUNCIL OF 5 JULY 2006 ON THE IMPLEMENTATION OF THE PRINCIPLE OF EQUAL OPPORTUNITIES AND EQUAL TREATMENT OF MEN AND WOMEN IN MATTERS OF EMPLOYMENT AND OCCUPATION (RECAST), *supra* note 46; COUNCIL DIRECTIVE 2004/113/EC OF 13 DECEMBER 2004 IMPLEMENTING THE PRINCIPLE OF EQUAL TREATMENT BETWEEN MEN AND WOMEN IN THE ACCESS TO AND SUPPLY OF GOODS AND SERVICES, *supra* note 47.

[102] Case C-13/94, P v S and Cornwall County Council, 1996 E.C.R. I-02143, http://curia.europa.eu/juris/liste.jsf?language=en&num=C-13/94.

[103] EUROPEAN UNION AGENCY FOR FUNDAMENTAL RIGHTS, *Protection against discrimination on grounds of sexual orientation, gender identity and sex characteristics in the EU – Comparative legal analysis – Update 2015* 7–8 (2015), https://fra.europa.eu/en/publication/2015/protection-against-discrimination-grounds-sexual-orientation-gender-identity-and (last visited Feb 25, 2020).

Identifying appropriate comparators is particularly difficult in cases of multi-dimensional discrimination. For example, it has been argued that a disabled woman must compare herself to a body-abled man.[105] A further well-known problem is age discrimination where a binary comparator cannot be identified.[106]

To address these barriers to identifying legitimate comparators EU non-discrimination law allows for "hypothetical comparators." A real comparator (i.e. a specific person or people that have received an advantage) thus does not necessarily need to be identified.[107] Hypothetical comparators are, for example, frequently seen as sufficient in pay equality cases (although these are mostly direct discrimination cases).[108]

The legitimacy of hypothetical comparators, however, varies across ECJ and Member State jurisprudence, particularly in relation to national origin as a proxy for ethnicity.[109] In *Maniero* the claimant alleged indirect discrimination under the Racial Equality Directive in relation to a rule that limited eligibility for a German scholarship to people that had passed the German Staatsexamen (a final exam for law students in German universities). The claimant, who completed his law degree in a third-party country, argued that this requirement disadvantaged people on the basis of ethnicity. The Court disagreed, concluding instead that disadvantaged group is not primarily made up by a particular ethnic group.[110] In order to establish disadvantage based on ethnicity, the

---

[104] FARKAS ET AL., *supra* note 7 at 42.

[105] MAKKONEN, *supra* note 38.

[106] CHRISTA TOBLER, *Limits and potential of the concept of indirect discrimination* 41 (2008) notes that "However, in some cases, it may be difficult to find comparators, for example in the case of age discrimination (Fredman 2003:56 et seq.; Hepple 2003:83; O'Cinneide 2005:26)."

[107] FARKAS ET AL., *supra* note 7 at 43. This could be, for example, an ideal minimum standard of treatment (e.g. human dignity). See also p. 45 on the arbitrary nature of differential treatment.

[108] McCrudden and Prechal, *supra* note 6 at 33.

[109] In this context it is important to note that the directive does not cover different treatment on grounds of nationality.

[110] CASE C-457/17, *supra* note 98 at 49 which states "in the present case, it is not disputed that the group to whom the Foundation grants an advantage as regards the award of the scholarships at issue in the main proceedings consists of persons who satisfy the requirement of having successfully completed the First State Law Examination, whereas the disadvantaged group consists of all persons who do not satisfy that requirement." The claimant attempted to compare their prior education to the German requirements.

Court concluded that "it is necessary to carry out, not a general abstract comparison," for example by equating national origin with ethnicity, but rather "a specific concrete comparison, in the light of the favourable treatment in question."[111]

The issue of whether a hypothetical or concrete comparator is necessary to establish disparity also emerged in the case *Jyske Finans*.[112] The claimant alleged that a Danish law requiring additional identification for citizens of third-party countries to obtain a loan to purchase a car causes indirect discrimination based on ethnic origin. The claimant argued that "persons of 'Danish ethnicity' will be treated more favourably as a result of the practice."[113] The ECJ did not agree. As in *Maniero* the Court ruled that "not a general abstract comparison, but a specific concrete comparison"[114] is needed to establish *prima facie* discrimination. Justifying this decision, the Court explained that "a person's country of birth cannot, in itself, justify a general presumption that that person is a member of a given ethnic group,"[115] and that it "cannot be presumed that each sovereign State has one, and only one, ethnic origin."[116] Thus, the Court found that because only one criterion (i.e. country of birth) was contested which is not synonymous with ethnicity, it cannot be assumed that the rule in question is discriminatory on ethnic grounds.[117] This ruling follows similar logic to the Court's judgements in *Jyske Finans* and *Chez*[118] where it was argued that the concept of ethnicity,[119] "[...] has its origin in the idea of societal groups marked in particular by common nationality, religious faith, language, cultural and traditional origins and backgrounds."[120]

---

[111] *Id.* at 48. Justifying this decision, the Court explains that "there is nothing in the documents before the Court to show that persons belonging to a given ethnic group would be more affected by the requirement relating to the First State Law Examination than those belonging to other ethnic groups." See: *Id.* at 50.

[112] Case C-668/15, Jyske Finans A/S v Ligebehandlingsnævnet, acting on behalf of Ismar Huskic, 2017 E.C.R. I–278, http://curia.europa.eu/juris/document/document.jsf?text=&docid=189652&pageIndex=0&doclang=EN&mode=lst&dir=&occ=first&part=1&cid=7165960 (last visited Aug 11, 2019).

[113] *Id.* at 28.

[114] *Id.* at 2.

[115] *Id.* at 20.

[116] *Id.* at 21. In this case the claimant would have needed to demonstrate that nationality is a reliable proxy for ethnicity.

[117] CHOPIN, CONTE, AND CHAMBRIER, *supra* note 87 at 17.

[118] Case C-83/14, CHEZ Razpredelenie Bulgaria AD v Komisia za zashtita ot diskriminatsi, 2015 E.C.R. I–480, http://curia.europa.eu/juris/document/document.jsf?docid=165912&doclang=EN.

[119] In general on the difficulties in defining the concept of "ethnic origin" see LILLA FARKAS ET AL., *The meaning of racial or ethnic origin in EU law: between stereotypes and identities.* (2017), http://bookshop.europa.eu/uri?target=EUB:NOTICE:DS0116914:EN:HTML (last visited Feb 9, 2020).

These rulings lie in stark contrast to a similar case heard by the Swedish Labour Court concerning a Swedish language requirement for employment. The Swedish court acknowledged that the requirement could result in *prima facie* indirect discrimination against anyone who is not Swedish (meaning a concrete comparator based on ethnicity was not needed), despite ultimately ruling against the claimant on the basis that the job requirement was justified.[121]

As these facets of the composition comparators demonstrate, the legitimacy of a proposed comparator depends on the circumstances of the case, arguments presented by claimants and alleged offenders, and the judgement of national and EU courts. Owing to the fact that they need to be transposed into national law, the non-discrimination directives themselves rarely offer clear cut requirements that do not require further judicial or regulatory interpretation.[122] Legitimacy is often not so much a legal question as a matter of skilled argumentation. National courts are granted a high margin of appreciation to interpret national legislation and the facts of the case.[123] As a result, debates over the legitimacy of a proposed comparator are often heated and opinions vary significantly across Member States.[124] Once again, the contextual determination of the legitimacy of a comparator poses a clear challenge for automating fairness at scale. It is not clear how such determinations can be legitimately made by system controllers on a case- or application-specific basis without judicial interpretation.

C.        REQUIREMENTS FOR A PARTICULAR DISADVANTAGE

Assuming a disadvantaged group and legitimate comparator can be identified, a 'particular disadvantage' suffered by a protected group must be demonstrated to establish *prima facie* discrimination. Conceptually, whether a harm can

---

[120] CASE C-668/15, *supra* note 115 at 17; CASE C-83/14, *supra* note 121 at 46.

[121] FARKAS AND O'DEMPSEY, *supra* note 7 at 37.

[122] In general on this issue see CHOPIN, CONTE, AND CHAMBRIER, *supra* note 87; Mark Bell, *The Implementation of European Anti-Discrimination Directives: Converging towards a Common Model?*, 79 POLIT. Q. 36–44 (2008); SUSANNE BURRI & HANNEKE VAN EIJKEN, GENDER EQUALITY LAW IN 33 EUROPEAN COUNTRIES: HOW ARE EU RULES TRANSPOSED INTO NATIONAL LAW? (2014); Susanne Burri & Aileen McColgan, *Sex Discrimination in the Access to and Supply of Goods and Services and the Transposition of Directive 2004/113/EC* 196.

[123] TOBLER, *supra* note 109 at 41; ELLIS AND WATSON, *supra* note 81; CASE C-1/95, *supra* note 73 at 35.

[124] McCrudden and Prechal, *supra* note 6 at 35–36.

be considered a 'particular disadvantage' depends upon its nature, severity, and significance:

- **Nature**: What is the harm and who does it affect?
- **Severity**: For each affected person, how severe or damaging is the harm?
- **Significance**: How many people from a protected group are disadvantaged, and how many in a comparable situation are advantaged?

Significance is more important than severity relatively speaking; a minor harm that affects many people can constitute a 'particular disadvantage' despite lacking in severity. As with considerations around the composition of the disadvantaged and comparator groups, consistent and explicit thresholds for each of these three elements do not exist in non-discrimination law and have not have generally not been advanced by the ECJ or national courts.

Concerning the nature of the harm, legal scholars, relevant legislation, and the ECJ differ about whether a concrete harm must have already occurred to establish a 'particular disadvantage'. For some, establishing a potential or hypothetical harm is sufficient.[125] Ellis and Watson, for example, suggests that the latter is sufficient as most legal systems aim to intervene if possible before harms actually occur.[126] This position is supported by EU racial discrimination directives which dictate that "indirect discrimination shall be taken to occur where an apparently neutral provision, criterion or practice would put persons of a racial or ethnic origin at a particular disadvantage compared with other persons  [emphasis added]."[127]

Support is also found in ECJ jurisprudence. Albeit a direct discrimination case, it was clear in *Feryn*[128] that hypothetical harm was sufficient to establish *prima facie* discrimination. Specifically, the Court ruled that the phrasing of a job advertisement indicating that immigrants would not be hired was likely to deter potential non-white applicants. The Court did not require proof of actual harm because the advertisement itself was seen as intuitively discriminatory. Similar reasoning is found in the Court's decision in *Accept* which addressed practices discouraging recruitment of gay football players.[129] Abstract harm was also sufficient in two cases concerning rights for part-time and full-time workers, where statistics concerning the gender of part-time workers were used to show that more women than men would stand to be affected by the contested rule.[130]

---

[125] ELLIS AND WATSON, *supra* note 81 at 152; Philipp Hacker, *Teaching fairness to artificial intelligence: Existing and novel strategies against algorithmic discrimination under EU law*, 55 COMMON MARK. LAW REV. 1143–1185 (2018).

[126] ELLIS AND WATSON, *supra* note 81 at 152.

[127] The same wording can be found in all four EU non-discrimination directives; see also *Id.* at 152.

[128] Case C-7/12, Nadežda Riežniece v Zemkopības ministrija and Lauku atbalsta dienests, 2013 ECLI:EU:C:2013:410, http://curia.europa.eu/juris/liste.jsf?num=C-7/12&language=EN.

The Court's position changes between cases. As discussed in relation to legitimate comparators (see: Section III.B), in *Maniero* unfavourable treatment needed to be shown not through "a general abstract comparison, but a specific concrete comparison, in the light of the favourable treatment in question."[131] Similar reasoning was used in *Jyske Finans*.[132]

While it remains open whether a hypothetical or abstract harm needs to be demonstrated, the Court's reasoning across the aforementioned cases would suggest that particular disadvantage must be established for specific practices or rules on a case-by-case basis. It seems unlikely that the Court would label all practices of a particular type (e.g. recruitment advertisements addressing protected characteristics) as necessarily *prima facie* discriminatory. If true, this type of case-specific judicial interpretation would be particularly difficult to automate, as classes of automated practices, rules, or types of applications could not be considered *prima facie* discriminatory without consulting the judiciary.

Concerning the severity and significance of the harm, whereas US anti-discrimination law utilises a rule of thumb to measure illegal disparity,[133] European jurisprudence is much more agile.[134] To cause a 'particular disadvantage' a concrete or hypothetical harm must be "particularly hard" for a specific individual or group. The jurisprudence is generally not concerned with a rule that "illustrates purely fortuitous or short-term phenomena,"[135] but rather a rule that "appears to be significant" in harming groups.[136] Similarly, if a rule similarly disadvantages everyone it can be deemed legal.[137] This intuitively makes sense.

---

[129] Case C-81/12, Asociația Accept v Consiliul Național pentru Combaterea Discriminării,, 2013 E.C.R. I–275, http://curia.europa.eu/juris/document/document.jsf?docid=136785&doclang=EN.

[130] Joined cases C-4/02 and C-5/02, Hilde Schönheit v Stadt Frankfurt am Main (C-4/02) and Silvia Becker v Land Hessen (C-5/02), 2003 E.C.R. I-12575, 63–64, https://eur-lex.europa.eu/legal-content/en/TXT/?uri=CELEX%3A62002CJ0004.

[131] CASE C-457/17, *supra* note 98 at 48, 50 in relation to goods and services (a university scholarship that required having passed the First State Law Examination). Here the Court ruled that "there is nothing in the documents before the Court to show that persons belonging to a given ethnic group would be more affected by the requirement relating to the First State Law Examination than those belonging to other ethnic groups".

[132] CASE C-668/15, *supra* note 115 at 31–32.

[133] Barocas and Selbst, *supra* note 5; Kim, *supra* note 5.

[134] Wachter, *supra* note 10; Catherine Barnard & Bob Hepple, *Indirect Discrimination: Interpreting Seymour-Smith*, 58 CAMB. LAW J. 399–412 (1999).

EU non-discrimination law and the ECJ have not established clear-cut thresholds for severity or significance for a 'particular disadvantage'. Thresholds are instead flexible and set on a case-by-case basis, if they are explicitly set by the judiciary at all.[138] More common is usage of imprecise phrases to indicate illegal disparity, such as "considerably more,"[139] "far more,"[140] "far greater number,"[141] "significantly high proportion of non-nationals, compared to nationals, are affected by that rule,"[142] "national measure, albeit formulated in neutral terms, works to the disadvantage of far more persons possessing the protected characteristic than persons not possessing it,"[143] "almost exclusively women," "significantly greater proportion of individuals of one sex as compared with individuals of the other sex,"[144] "affects considerably more women than men,"[145] "much larger number of women than of men,"[146] "considerably lower percentage,"[147] "a considerably lower percentage of men,"[148] "in percentage terms considerably less women,"[149] "considerably higher number of women than men,"[150] "a considerably higher percentage of women,"[151] "affected considerably," "much larger number of women than of men,"[152] "a greater number of women than men,"[153] and "more women."[154] Concerning the advantaged group illegal disparity occurs when a "considerably smaller percentage of women"[155] or "considerably smaller proportion of women than men"[156] are able to satisfy a contested rule than others in a similar situation.

---

[135] CASE C-167/97, *supra* note 71 at 62; TOBLER, *supra* note 109 at 41.

[136] CASE C-167/97, *supra* note 71 at 62 as well as the case law cited within.

[137] MAKKONEN, *supra* note 38 at 37.

[138] EUROPEAN UNION AGENCY FOR FUNDAMENTAL RIGHTS AND COUNCIL OF EUROPE, HANDBOOK ON EUROPEAN NON-DISCRIMINATION LAW 242–243 (2018 edition ed. 2018), https://fra.europa.eu/sites/default/files/fra_uploads/1510-fra-case-law-handbook_en.pdf.

[139] Case C-363/12, Z. v A Government department, The Board of management of a community school, 2014 E.C.R. I–159, 53, http://curia.europa.eu/juris/liste.jsf?language=en&num=C-363/12 In this case the Court held that "[t]he Court has consistently held that indirect discrimination on grounds of sex arises where a national measure, albeit formulated in neutral terms, puts considerably more workers of one sex at a disadvantage than the other." The Court also cited the following cases in support; CASE C-1/95, *supra* note 73 at 30; Case C-123/10, Waltraud Brachner v Pensionsversicherungsanstalt, 2011 E.C.R. I-10003, 56, http://curia.europa.eu/juris/liste.jsf?language=en&num=C-123/10; CASE C-7/12, *supra* note 131 at 39.

[140] Case C-527/13, Lourdes Cachaldora Fernández v Instituto Nacional de la Seguridad Social (INSS) and Tesorería General de la Seguridad Social (TGSS), 2015 ECLI:EU:C:2015:215, 28, https://eur-lex.europa.eu/legal-content/EN/TXT/?uri=CELEX%3A62013CJ0527; CASE C-123/10, *supra* note 142 at 56 and the case law cited; Case C-385/11, Isabel Elbal Moreno v Instituto Nacional de la Seguridad Social (INSS), Tesorería General de la Seguridad Social (TGSS), 2012 EU:C:2012:746, 29, http://curia.europa.eu/juris/document/document.jsf?text=&docid=185565&pageIndex=0&doclang=EN&mode=lst&dir=&occ=first&part=1&cid=7600685.

[141] CASE C-170/84, *supra* note 68; CASE C-256/01, *supra* note 65; CASE 171/88, *supra* note 69.

[142] CASE C-703/17 OPINION OF ADVOCATE GENERAL LEGER, *supra* note 98 at 61 in relation to the acknowledgement of prior employment years for higher salaries in a different country which was not seen as comparable.

[143] CASE C-668/15, *supra* note 115 at 30 in relation to the need for third party residents to offer additional means of identification which was not seen as discriminatory; CASE C-83/14, *supra* note 121 at 101.

As these phrases suggest, specific percentages or thresholds for significant disparity are rarely set by the judiciary. There are, however, exceptions. In *Moreno* the Court argued that an adverse action constitutes discrimination if 80% of the affected group are women.[157] In two other cases relating to different rights for part-time workers compared to full-time workers, *prima facie* discrimination was established in cases where 87.9%[158] and 87%[159] of part-time

---

employees were women. Further, in his opinion for *Nolte,* AG Léger suggested that:

> "[I]n order to be presumed discriminatory, the measure must affect 'a far greater number of women than men' or 'a considerably lower percentage of men than women' or 'far more women than men'…Consequently, the proportion of women affected by the measure must be particularly marked. Thus, in the judgement *Rinner-Kühn*, the Court inferred the existence of a discriminatory situation where the percentage of women was 89%. In this instance, *per se* the figure of 60 % [...] would therefore probably be quite insufficient to infer the existence of discrimination."[160]

The threshold of significant or severe disparity can be lowered if evidence "revealed a persistent and relatively constant disparity over a long period."[161] In cases of the European Court of Human Rights the threshold has also been lowered when the affected party is part of a minority group (e.g. Roma).[162]

The timing of discrimination must also be considered. The Court has acknowledged that various points in time can be assessed to establish whether discrimination occurred.[163] For example, a law could be assessed at the time when it was enacted or when the discrimination actually occurred. This is a sensible approach as it acknowledges that inequalities can change over time as society changes.[164] A law that disadvantages part-time workers might be disproportionately burdensome on women when it is enacted, but less so over time if the ratio of women to men in part-time work balances out over time. It is ultimately for national courts to decide on the appropriate timing for assessing disparity and whether the relevant "statistical evidence is valid, representative and significant."[165] Yet again the Court's determination of an essential component of discriminatory is reliant upon imprecise and flexible concepts which cannot easily be replicated in automated systems without appeal to case- or application-specific judicial interpretation.

---

[160] Case C-137/93 Opinion of Advocate General Leger, Inge Nolte v. Landesversicherungsanstalt Hannover, 1995 E.C.R. I–438, 57–58, http://curia.europa.eu/juris/showPdf.jsf?text=&docid=99269&doclang=EN (last visited Mar 26, 2019).

[161] CASE C-167/97, *supra* note 71. para

[162] Case 57325/00 ECHR, D.H. and Others v. the Czech Republic, 2007, https://hudoc.echr.coe.int/eng?i=001-83256. Standardised testing has a negative effect on children (i.e. being placed in special schools) and can significantly impact a particular minority if the class is composed of 50-90% Roma children. This is seen as discriminatory due to Roma people only making up 2% of the general population.

[163] CASE C-167/97, *supra* note 71 at 48–49.

[164] CASE C-167/97, *supra* note 71.

[165] CASE C-161/18, *supra* note 71 at 45.

D.    ADMISSIBILITY AND RELEVANCE OF EVIDENCE

The final requirement is to provide convincing and relevant evidence that a protected group has suffered a 'particular disadvantage' in comparison with people in a similar situation. At least four types of evidence can be submitted to establish that a particular disadvantage has occurred: (1) statistical evidence of significant disparity[166]; (2) "common sense assessment" (or a liability test)[167]; (3) situation testing (although its admissibility is contested)[168]; and (4) inferences drawn from circumstantial evidence.[169] Any of the four types of evidence can be sufficient to establish a particular disadvantage and *prima facie* discrimination.

   *1.    Statistical evidence*

EU jurisprudence shows that statistical evidence has been infrequently used to establish *prima facie* discrimination (UK case law being an exception).[170] Cases in which statistical evidence has been used tend to address claims relating to unequal pay based on sex, redundancy based on age, and racial segregation.[171]

Compared with direct discrimination where the claimant only needs to prove that they themselves were treated less favourably based on a protected attribute, indirect discrimination invites greater use of statistics because protected attributes are, by definition, not explicitly used.[172] Statistical evidence can be indispensable to establish a possible or actual correlation between legally protected attributes and the factors considered in the contested rule. Of course, statistics can also be helpful in direct discrimination cases if the claimant wants to prove a certain illegal pattern, for example that a company does not hire people of a particular ethnicity despite much of the population belonging to the group in question.[173]

---

[166] CASE C-167/97, *supra* note 71.
[167] TOBLER, *supra* note 109 at 40; MAKKONEN, *supra* note 38 at 34.
[168] Case C-423/15, Nils-Johannes Kratzer v R+V Allgemeine Versicherung AG, 2016 ECLI:EU:C:2016:604, https://eur-lex.europa.eu/legal-content/EN/TXT/?uri=CELEX%3A62015CJ0423; see also ELLIS AND WATSON, *supra* note 81; EUROPEAN UNION AGENCY FOR FUNDAMENTAL RIGHTS AND COUNCIL OF EUROPE, *supra* note 141 at 241 explains that "it seems that the CJEU adopted a different approach to 'situation testing'. Example: The case of Nils-Johannes Kratzer v. R+V Allgemeine Versicherung AG680 concerns a lawyer who had applied for a job solely to bring a discrimination complaint rather than with a view to obtaining that position." This was also seen as a potential misuse of abuse of rights. see also FARKAS ET AL., *supra* note 7 at 24.
[169] FARKAS AND O'DEMPSEY, *supra* note 7 at 47 argue that there are a range of other tools such as "questionnaires, audio or video recording, forensic expert opinion and inferences drawn from circumstantial evidence"; see also FARKAS ET AL., *supra* note 7 at 36.
[170] MAKKONEN, *supra* note 38 at 30.
[171] FARKAS AND O'DEMPSEY, *supra* note 7 at 49.
[172] MAKKONEN, *supra* note 38 at 31.

The relative lack of statistical evidence in prior jurisprudence reflects a general reluctance among legal scholars to require statistical evidence to establish *prima facie* discrimination.[174] There are sound reasons for this attitude. Advocate General Ledger explicitly mentions that requiring statistics could result in a "battle of numbers"[175] that implicitly favours claimants or offenders capable of producing convincing statistics. Similarly, Barnard and Hepple warn of "faulty statistical and factual assumptions"[176] and explain that "[i]n order to be reliable sophisticated statistical techniques are required which lie beyond the resources of parties in individual legal suits."[177] At the same time, it can be difficult for courts to ignore statistics if they are available or submitted by claimants or offenders.[178] With that said, the judiciary is ultimately empowered to determine the admissibility of statistics on a case-by-case basis; numerous examples exist of convincing and relevant statistics being ignored as incidental evidence (e.g. the law of women in the workforce and in management positions).[179]

Prior jurisprudence shows that the composition of the disadvantaged group influences the admissibility of evidence to establish *prima facie* and actual discrimination on a case-by-case basis. Comparative evidence must reflect the geographic and demographic reach of the contested rule, which in turn influences who is seen as a legitimate comparator. Based on the reach[180] of the rule only statistical evidence describing the general population of the Member State or region, or a particular company, sector, trade or other social or professional grouping may be seen as relevant and admissible. The contested rule could, for

---

example, be an individual contract,[181] legislation,[182] or collective agreement[183] and thus legitimate statistics will vary.

Several prior cases show how this dependence between reach and the relevance of statistical evidence works in practice. The ECJ has previously rejected statistical evidence showing that part-time work in Berlin is primarily carried out by women on the basis that the contested rule was a regulatory provision applicable to all of Germany, not only Berlin.[184] As a result the alleged disadvantage would have needed to be established at a population level rather than a regional level. The ECJ has similarly rejected global statistics in cases where Spanish legislation was contested.[185] In contrast, a national court in Germany allowed statistics about the population of Bavaria as the contested rule was a law applicable to civil servants only in this region.[186]

Overreliance on statistics can also undermine efforts to establish equality in areas where relevant statistics do not exist, but where potential discrimination is "fairly obvious as a matter of common sense."[187] Statistics regarding sensitive issues, for example in relation to disability, sexual orientation, or ethnic origin are often not available.[188] Personal data of other employees in employment discrimination cases, for example, are often unavailable to claimants which has posed a barrier to these cases in the past.[189] This is unsurprising and intentional in part, as European privacy and data protection law aim to mini-mise the collection and storage of sensitive personal data.[190]

---

[181] CASE C-170/84, *supra* note 68.

[182] CASE 171/88, *supra* note 69.

[183] CASE C-127/92, *supra* note 67.

[184] CASE C-300/06, *supra* note 73.

[185] CASE C-527/13, *supra* note 143.

[186] CASE C-1/95, *supra* note 73.

[187] ELLIS AND WATSON, *supra* note 81 at 151. The lack of statistics concerning traditionally marginalised groups suggests that introducing new requirements for statistical evidence to enable consistent identification and assessment of automated discrimination risks further marginalising these groups. With that said, historical reasons exist in Europe to not push for further collection of sensitive personal data from minority and marginalised groups. See: Footnote 190.

[188] *Id.* at 151, 155. referring to "statistics are unlikely to be obtainable" and "ethnic data where there are sensitivities"; TOBLER, *supra* note 109 at 41; FARKAS ET AL., *supra* note 122 at 116.

[189] Case C-104/10, Patrick Kelly v National University of Ireland (University College, Dublin), 2011 E.C.R. I-06813, http://curia.europa.eu/juris/liste.jsf?num=C-104/10; Case C-415/10, Galina Meister v Speech Design Carrier Systems GmbH, 2012 ECLI:EU:C:2012:217, http://curia.europa.eu/juris/liste.jsf?num=C-415/10 Both cases dealt with discrimination in the workplace. The Court ruled that the employer is not required to release personal data of other employees in order to facilitate actions brought by the claimant. In both cases data protection law trumped the claimant's need for the data to prove their case.

Nonetheless, many legal scholars, policy bodies, and researchers recognise the potential value of statistical evidence to detect, mitigate, and adjudicate discrimination. The ECJ, for example, has recognised the value of statistics to refute alleged discrimination (see: Section F).[191] The algorithmic fairness community in particular has developed methods which require special category data to detect and mitigate biases in training data and automated decisions.[192] These communities are understandably increasingly calling for greater collection of special category data to facilitate discrimination detection and legal proceedings.[193] Some policy bodies such as the European Committee of Social Rights and the United Nations Special Effort on Extreme Poverty and Human Rights have advocated for an ethical responsibility to collect sensitive data to facilitate legal proceedings.[194] At the same time, simply collecting more sensitive data will not solve the problems created by new automated technologies; mitigation strategies, effective remedies and assessment procedures must also be in place.[195]

---

[190] EUROPEAN COMMISSION, *Regulation (EU) 2016/679 of the European Parliament and of the Council of 27 April 2016 on the protection of natural persons with regard to the processing of personal data and on the free movement of such data, and repealing Directive 95/46/EC (General Data Protection Regulation)* (2016), http://eur-lex.europa.eu/legal-content/EN/TXT/PDF/?uri=OJ:L:2016:119:FULL&from=EN Article 5; scholarship linking data privacy and non-discrimination law see Ignacio Cofone, *Antidiscriminatory Privacy*, SSRN ELECTRON. J. (2018), https://www.ssrn.com/abstract=3154518 (last visited Feb 9, 2020); Hacker, *supra* note 128. This rationale is strongly influenced by European history. European data protection and privacy laws in Europe are also a response to the experiences of the Second World War, which showed dramatically how sensitive information can be misused with catastrophic consequences. For further commentary on the history of privacy and data protection in Europe see: CHRISTOPH GRABENWARTER, THE EUROPEAN CONVENTION FOR THE PROTECTION OF HUMAN RIGHTS AND FUNDAMENTAL FREEDOMS: A COMMENTARY (01 edition ed. 2014).

[191] CASE C-7/12, *supra* note 131; CASE C-81/12, *supra* note 132.

[192] Cynthia Dwork et al., *Fairness through awareness*, *in* PROCEEDINGS OF THE 3RD INNOVATIONS IN THEORETICAL COMPUTER SCIENCE CONFERENCE 214–226 (2012); Cynthia Dwork & Deirdre K. Mulligan, *It's not privacy, and it's not fair*, 66 STAN REV ONLINE 35 (2013).

[193] MAKKONEN, *supra* note 38 at 18, 25; FARKAS ET AL., *supra* note 7 at 82.

[194] FARKAS ET AL., *supra* note 122 at 118.

[195] SANDRA FREDMAN, EUROPEAN COMMISSION & EUROPEAN NETWORK OF LEGAL EXPERTS IN THE FIELD OF GENDER EQUALITY, *Making Equality Effective: The role of proactive measures* 30–31 (2009).

### 2. Intuitive and traditional evidence

Regardless of how this debate evolves going forward its mere existence indicates that the availability of relevant, accurate, and impartial statistics thus cannot be taken for granted. As a result, while admissible, the legal community and judiciary have been reluctant to require statistical evidence to establish *prima facie* discrimination.[196] Many scholars see value in common sense assessments according to which a contested rule need only be judged as intuitively potentially discriminatory, or 'on the face' discriminatory.[197] Tolber[198] explains that this approach allows "common knowledge,"[199] "obvious facts,"[200] or "convictions"[201] to be taken into consideration. Examples could include how a ban on turbans or beards at work could, for example, have an indirect discriminatory effect on Sikh people or people of Pakistani origin.[202] Similarly, the Dutch Equal Treatment Commission has suggested that an imbalance in the composition of the workforce in relation to ethnicity can itself be a strong indication of indirect discrimination.[203]

The ECJ at times has held a similar view. In *Meister* the Court stated that "indirect discrimination may be established by any means, and not only on the basis of statistical evidence."[204] This view is consistent with the case law of the Member States where actual disadvantage does not to be proven, but a "typical tendency"[205] is sufficient (see: Section C). Elsewhere, in *Schnorbus* the Court held that a rule giving preferential treatment to people with prior military service when applying for jobs was self-evidently discriminatory for women because only men are required to serve in the military in Germany. In this case statistical evidence regarding gender distribution in the military or the general population was not required as the underlying intuition that such a policy could result in gender-based discrimination is obviously reasonable.[206]

---

[196] Barnard and Hepple, *supra* note 137; ELLIS AND WATSON, *supra* note 81.

[197] TOBLER, *supra* note 109 at 40; MAKKONEN, *supra* note 38 at 34; Oran Doyle, *Direct discrimination, indirect discrimination and autonomy*, 27 OXF. J. LEG. STUD. 537–553, 540 (2007).

[198] For all three see TOBLER, *supra* note 109 at 40.

[199] ELLIS AND WATSON, *supra* note 81 at 151 which discusses "potentially discriminatory" legislation; Case C-322/98, Bärbel Kachelmann v Bankhaus Hermann Lampe KG, 2000 E.C.R. I-07505, 24, http://curia.europa.eu/juris/liste.jsf?language=en&num=C-322/98 noting "it is common ground in Germany."

[200] Case C-79/99, Julia Schnorbus v Land Hessen, 2000 E.C.R. I-10997, https://eur-lex.europa.eu/legal-content/EN/TXT/?uri=CELEX%3A61999CJ0079.

[201] Case C-237/94, John O'Flynn v Adjudication Officer, 1996 E.C.R. I-02617, https://eur-lex.europa.eu/legal-content/EN/TXT/?uri=CELEX%3A61994CJ0237.

[202] Examples taken from UK case law see MAKKONEN, *supra* note 38 at 34; ELLIS AND WATSON, *supra* note 81 at 168.

[203] MAKKONEN, *supra* note 38 at 34.

[204] CASE C-161/18, *supra* note 71 at 46 referring to; CASE C-415/10, *supra* note 192 at 43.

[205] FARKAS AND O'DEMPSEY, *supra* note 7 at 50.

This is not to suggest that alternatives to statistical evidence are inherently preferable; rather, the relevance of different types of evidence depends on the facts of the case. 'Traditional' types of evidence, such as documentary evidence, witness statements, or expert opinions are often ineffective in cases of indirect discrimination,[207] especially when a correlation between a protected attribute and non-protected attribute needs to be demonstrated. Non-statistical evidence is also not necessarily easier to obtain than statistical evidence. Nonetheless, given that automated discrimination may only be observable at a statistical level, statistical evidence may need to become a default option in the future.

In general, the ECJ has not established consistent requirements to evaluate the reliability, relevance and significance of statistical and other types of evidence.[208] Setting these requirements is often delegated to national courts which has led to an inconsistent evidential standard across Europe.[209] National courts enjoy a high margin of appreciation to interpret non-discrimination law as they see fit.[210] In most cases regional courts, because they are the closest authority to the case, will be the most competent authority to assess whether contested rules affect protected groups, the extent to which harm occurred, and which evidence is reliable and significant to establish, disprove, or justify a particular disadvantage (see: Section F). This delegation of authority is an intentional feature of the law which we refer to as 'contextual equality'. This contextual and inconsistent approach to assessing which evidence should be admitted in court, whilst sensible and desirable in terms of the rule of law, makes automating fairness particularly challenging.

E.        A COMPARATIVE 'GOLD STANDARD' TO ASSESS DISPARITY

Once evidence of a particular disadvantage has been offered, and the composition of the disadvantaged and comparator groups has been determined, the judiciary must decide whether the disadvantaged group has been or are likely to be significantly harmed in comparison to another group in a similar situation, and thus whether *prima facie* discrimination exists. The jurisprudence is inconsistent concerning how to measure legal inequality between groups.

---

[206] CASE C-79/99, *supra* note 203 at 38.
[207] FARKAS ET AL., *supra* note 7 at 24.
[208] TOBLER, *supra* note 109 at 41 as well as the case law cited within.
[209] CASE C-161/18, *supra* note 71 at 45.
[210] TOBLER, *supra* note 109 at 41; ELLIS AND WATSON, *supra* note 81; CASE C-1/95, *supra* note 73 at 35; CASE C-170/84, *supra* note 68.

The Court has three options: examine the effect of a contested rule on (1) only the disadvantaged group, (2) only the comparator (advantaged) group, or (3) both the disadvantaged and comparator groups.

Interestingly, legal scholars and the wording of directives concerning gender and racial inequality suggest that only the disadvantaged group needs to be evaluated to determine whether a particular disadvantage occurred.[211] This is especially problematic for racial inequality in cases where the protected group is not evenly split across the population.

Interestingly, only requiring examination of the disadvantaged group stands at odds with a 'gold standard' for comparing groups in non-discrimination cases advanced by the ECJ. In *Seymour-Smith* the Court argued that a full comparison is the "best approach" to assessing potentially discriminatory disparity:

> "the best approach to the comparison of statistics is to consider, on the one hand, the respective proportions of men in the workforce able to satisfy the requirement of two years' employment under the disputed rule and of those unable to do so, and, on the other, to compare those proportions as regards women in the workforce. It is not sufficient to consider the number of persons affected, since that depends on the number of working people in the Member State as a whole as well as the percentages of men and women employed in that State."[212]

This view has subsequently been re-affirmed in multiple cases.[213] Despite the Court having described a 'gold standard' for comparison,[214] this has not led to consistent comparison of disparity by the ECJ and national courts. In many cases the court has examined only the disadvantaged group. In *Gerster* the Court, in examining discrimination in workplace promotion policies, looked only at the disadvantaged group despite statistics about the advantaged group being available.[215] In this case, part-time workers that worked between one-half and two-thirds of a full-time workload had their length of service reduced to two-thirds of the number of years employed. In assessing whether the promotion policy contravened Directive 76/207/EEC on gender discrimination in employment,[216] the Court only looked at the fact that 87% of the part-time workers in the Bavarian civil service were women.[217] On this basis the promotion policy was found to be discriminatory barring any further objective justification of the practice by the alleged offender (see: Section F).[218] Similarly in *Bilka*, which addressed part-time workers being excluded from pension schemes, the Court only looked at the disadvantaged group and ignored the advantaged group, despite the accused company having provided statistics showing that pensions were paid to women at more than a 1:1 ratio: 81.3% of the pensions went to women despite the workforce consisting of only 72% women.[219]

---

[211] TOBLER, *supra* note 109 at 41; SANDRA FREDMAN, DISCRIMINATION LAW 111 (2002); L. A. J. Senden, *Conceptual convergence and judicial cooperation in sex equality law*, *in* THE COHERENCE OF EU LAW: THE SEARCH FOR UNITY IN DIVERGENT CONCEPTS 363–396, 374 (2008).

[212] CASE C-167/97, *supra* note 71 at 59.

[213] CASE C-300/06, *supra* note 73 at 41; CASE C-161/18, *supra* note 71 at 39; CASE C-7/12, *supra* note 131; CASE C-81/12, *supra* note 132; for an overview of relevant jurisprudence up to 2008 see: TOBLER, *supra* note 109 at 41.

[214] TOBLER, *supra* note 109 at 41; FREDMAN, *supra* note 214 at 111; Senden, *supra* note 214 at 374.

[215] CASE C-1/95, *supra* note 73.

Elsewhere, the Court has only explicitly examined the advantaged group. Tobler explains that in sex discrimination cases the Court has often not followed its own standard.[220] Even in *Seymour-Smith*, where the 'gold standard' was first proposed, the Court opted to explicitly look only at the advantage group. Specifically, to support their ruling that a two-year employment requirement necessary to raise unfair dismissal claims with the UK Industrial Tribunal did not constitute indirect discrimination on the basis of gender, the Court cited statistics showing that 77.4% of men and 68.9% of women in the workforce were able to meet the requirement. The requirement was therefore not seen to cause indirect discrimination because of the relatively small difference in the percentage of qualifying men and women.[221] Elsewhere, in *Maniero* the Court compared both groups (people eligible to apply for a scholarship in Germany), but did not assess the actual composition of the groups in terms of ethnicity.[222]

---

[216] EUROPEAN COUNCIL, *supra* note 93 at 76.

[217] CASE C-1/95, *supra* note 73 at 33.

[218] *Id.* at 42.

[219] CASE C-170/84, *supra* note 68 at para 7.

[220] TOBLER, *supra* note 109 at 41.

[221] Based on the statistics cited by the Court, it could be argued that the advantaged group was implicitly compared to the disadvantaged group. If 77.4% of men and 68.9% of women qualified (i.e. the advantaged group), we can logically deduce that 22.6% of men and 31.1% of women did not qualify (i.e. the disadvantaged group). While true, this argument misses that the Court did not explicitly address the size of both groups according to gender in their ruling. Particularly for cases not dealing with binary characteristics (e.g. 'men' and 'women' when sex is the relevant characteristic), the type of implicit comparison between groups undertake by the Court can prove problematic. We examine the importance of comparing disadvantaged and advantaged groups both in terms of percentages and quantity in Sections III.E and VI.B. However, as we will show in Section V.A *Id.* at 41 and other scholars believe that even when a contested rule is based on ethnicity (i.e. where a 50:50 split in population between protected groups cannot be assumed) the law only requires examination of the disadvantaged group.

Whilst it makes intuitive sense to start by examining the composition of only the disadvantaged or advantaged group, it is crucial to consistently assess both groups because certain patterns of inequality might only emerge through direct comparison.[223] The Court recognised this much in *Seymour-Smith*,[224] *Voss*,[225] *Maniero*,[226] and most recently in *Violeta Villar Láiz*.[227] As mentioned above, if Danish[228] and German[229] citizens were primarily treated more favourably than other people in a comparable situation in these cases, the fact that the disadvantaged group is not primarily composed of a specific ethnicity does not mean that the contested rule is *prima facie* legitimate. This remains true regardless of how the social construct of ethnicity is defined and regardless of the ECJ's ruling in these cases. Further, it is especially true in cases where the protected group is not evenly split across the general population (e.g. ethnicity, disability, religion, sexual orientation).

The judicial interpretations of ethnicity across these cases are particularly problematic in cases of intersectional discrimination (see: Section A.1), as well as in cases where the majority of the disadvantaged group is not made up of a single ethnic group but rather a mix of people (e.g. hiring policies that prefer white men over anyone else). If groups can only be defined by a single protected trait, a rule that disadvantages at an equal rate across protected groups could give the impression that all groups have been treated fairly while ignoring the greater disparity experienced by individuals belonging to more than one group.[230] Comparison with the advantaged group(s) is a minimal requirement to reveal intersectional or additive disparity. This can likewise limit possible disparity to majority groups (see: Section V.B).

---

[222] CASE C-457/17, *supra* note 98 at 49 which states "in the present case, it is not disputed that the group to whom the Foundation grants an advantage as regards the award of the scholarships at issue in the main proceedings consists of persons who satisfy the requirement of having successfully completed the First State Law Examination, whereas the disadvantaged group consists of all persons who do not satisfy that requirement." The claimant attempted to compare their prior education to the German requirements.

[223] In real-life cases such as the impact promotion practices on part-time work, it was intuitively clear that these will affect women significantly more than men due to the knowledge that historically, men predominantly carried out full-time work. In the same way it was safe to assume that the statistical distribution of the general population will be reflected inside the company. Examination of the disadvantaged group only was therefore sufficient because of this common knowledge. These rules of thumb will not necessarily hold in the future as social roles and demographics change. See: CASE C-167/97, *supra* note 71; See also UK case law examples in ELLIS AND WATSON, *supra* note 81; See also CASE C-300/06, *supra* note 73; CASE C-1/95, *supra* note 73 regarding the community as a whole.

[224] CASE C-167/97, *supra* note 71 at 59.

[225] CASE C-300/06, *supra* note 73 at 41.

[226] CASE C-457/17, *supra* note 98 at 49.

[227] CASE C-161/18, *supra* note 71 at 39 in relation to part-time work.

[228] CASE C-668/15, *supra* note 115.

[229] CASE C-457/17, *supra* note 98.

From all of the considerations above it follows that only a comparison with the composition of the advantaged group can reveal that a homogenous group is favoured at the expense of a heterogenous group. Consider for example a hypothetical company that rejects candidates at equal rates across ethnicity and gender. Looking solely at the disadvantaged group (i.e. people who were not hired) would make the firm's hiring practices appear *prima facie* non-discriminatory. However, examination of the advantaged group could reveal that only white men were being hired. Only by examining both groups can the nature and magnitude of disparity caused by a given rule be understood in full (see: Section VI).

Across these cases the ECJ and national courts have established a "flexible and pragmatic" test for comparing effects on the disadvantaged and comparator group.[231] To meet a European legal standard any attempt to automate judgements of discrimination would need to accommodate this flexibility and ideally facilitate the judiciary's 'gold standard' for full comparisons between disadvantaged and advantaged groups.

F.      REFUTING *PRIMA FACIE* DISCRIMINATION

Direct discrimination is only lawful if it is explicitly legally allowed, for example on the basis of a "genuine occupational requirement."[232] Indirect discrimination can be justified if a legitimate aim is pursued and the measure is necessary and proportionate. The following focuses primarily on justification in cases of indirect discrimination which will likely make up the bulk of cases of automated discrimination (see: Section II). With that said, some considerations are also relevant for direct discrimination cases.

Assuming *prima facie* discrimination has been established, the burden of proof shifts from the claimant to the alleged offender who can then try to refute the alleged discriminatory rule.[233] This shift can occur when a claimant has made a convincing case for *prima facie* discrimination, or when "the causation between the protected ground and the harm is only probable or likely."[234] The shift can also occur if an alleged offender refuses to offer relevant information necessary for claimants to make their case; such a refusal can itself be grounds to establish *prima facie* discrimination.[235] If alleged offenders refuse to submit responses (e.g. via questionnaires), this lack of transparency can justify the court shifting the burden of proof, thereby forcing the alleged offender to respond to the claims of discrimination.[236] Ultimately, it is left to the national authorities and Member state law to determine the point at which the burden of proof shifts.[237]

---

[230] See for example: CASE C-443/15, *supra* note 83.

[231] TOBLER, *supra* note 109 at 41; MAKKONEN, *supra* note 38 at 36.

[232] For example Art 4 of COUNCIL DIRECTIVE 2000/78/EC OF 27 NOVEMBER 2000 ESTABLISHING A GENERAL FRAMEWORK FOR EQUAL TREATMENT IN EMPLOYMENT AND OCCUPATION, *supra* note 48.

[233] FARKAS ET AL., *supra* note 7; CASE C-167/97, *supra* note 71 at 57.

Assuming this shift occurs, in general there are two ways in which the alleged offender can be freed from liability for indirect discrimination: (1) by refuting that a causal link between the differential results and a protected ground does not exist,[238] or (2) by acknowledging that differential results have occurred but providing a justification that is based on the pursuit of a legitimate interest in a necessary and proportionate manner.[239] Both parties are free to submit any statistical evidence they believe to be relevant and significant to support, refute, or justify such claims.[240] Statistical evidence can be particularly valuable for disputing an alleged causal link between indirect discrimination and protected grounds.[241]

---

[234] Julie Ringelheim, *The Burden of Proof in Antidiscrimination Proceedings. A Focus on Belgium, France and Ireland*, FOCUS BELG. FR. IREL. SEPT. 4 2019 EUR. EQUAL. LAW REV., 7 (2019).

[235] FARKAS ET AL., *supra* note 7 at 9; Case 109/88, Handels- og Kontorfunktionærernes Forbund I Danmark v Dansk Arbejdsgiverforening, acting on behalf of Danfoss, 1989 E.C.R. I-03199, https://eur-lex.europa.eu/legal-content/EN/TXT/?uri=CELEX%3A61988CJ0109.

[236] FARKAS ET AL., *supra* note 7 at 37.

[237] *Id.* at 54.

[238] This defence is also valid in direct discrimination cases.

[239] CASE C-170/84, *supra* note 68 in para 45 where it states that this is the case when "the means chosen for achieving that objective correspond to a real need on the part of the undertaking , are appropriate with a view to achieving the objective in question and are necessary to that end." In direct discrimination cases only a legal provision could justify the different treatment.

[240] On how no coherent standards for admissable statiscs exist see FARKAS AND O'DEMPSEY, *supra* note 7 at 49.

[241] A less 'data intensive' and more argumentative approach to refute an alleged causal link can be pursued to show that the advantaged and disadvantaged groups are not in a comparable situation. As mentioned in Section III.C examples include that workers and employees are not seen being in a comparable situation because workers on average are less educated or have lower employment requirements than employees.

While the wording of the law (and in some cases, the judiciary) does not focus on the advantaged group, alleged offenders sometimes use this group to refute discrimination claims.[242] Examination of the advantaged group can help disprove a causal link between differential results and the protected grounds. For example, in *Bilka* the alleged offending company provided statistics showing that 81.3% of pensions are paid to women, despite women only composing 72% of their employees.[243] The aim of submitting these statistics was to refute the claim that excluding part-time workers from pension schemes disproportionately affects women. Similarly, in cases of alleged age discrimination concerning redundancy, a government agency has provided evidence to show that workers over the age of 50 were still employed at the firm.[244]

Looking at the advantaged group to refute alleged discrimination is also something that the Court has encouraged. In *Feryn* and *Accept*, both of which were direct discrimination cases, examination of the advantaged group was recommended as potential means to free oneself from liability. In *Feryn*[245] an employer advertised a position stating that "immigrants" will not be hired. Similarly, in *Accept*[246] a representative of a football club stated that he would not hire gay players. In both cases the Court explained that these claims could be refuted by referring to inclusive hiring polices (e.g. showing that immigrants and gay people are routinely recruited and thus part of the advantaged group).[247]

The second defence available to alleged offenders is to acknowledge indirect discrimination but offer a justification by claiming that the contested rule pursues a legitimate interest in a necessary and proportionate manner.[248] Of course, what constitutes a legitimate interest[249] and which measures are acceptable is, like many other determinations in non-discrimination cases, dependent upon the context of the case and the interpretation of national courts and relevant legislation. The ECJ gives a high margin of appreciation to the national courts to interpret national legislation and the facts of the case,[250] albeit with some limitations.[251]

---

[242] MAKKONEN, *supra* note 38 at 33.

[243] CASE C-170/84, *supra* note 68 at para 7.

[244] It should be noted that this relates to a case heard my the Danish Supreme Court EUROPEAN UNION AGENCY FOR FUNDAMENTAL RIGHTS AND COUNCIL OF EUROPE, *supra* note 141 at 243; on the legal age of discrimination see ELLIS AND WATSON, *supra* note 81 at 408–418.

[245] CASE C-7/12, *supra* note 131.

[246] CASE C-81/12, *supra* note 132.

[247] CASE C-7/12, *supra* note 131; CASE C-81/12, *supra* note 132.

[248] CASE C-170/84, *supra* note 68.

[249] See Justyna Maliszewska-Nienartowicz, *Direct and indirect discrimination in European union law–how to draw a dividing line*, 3 INT. J. SOC. SCI. 41–55, 44 (2014) who states that "'ensuring coherence of the tax system; the safety of navigation, national transport policy and environmental protection in the transport sector; protection of ethnic and cultural minorities living in a particular region; ensuring sound management of public expenditure on specialised medical care; encouragement of employment and recruitment by the Member States;

The need for contextual assessment of legitimate interests and justifications offered by alleged offenders means adds an additional layer of complexity to judicial interpretation. The alleged offender is free to contest any of the arguments and evidence offered by a claimant concerning the reach of the contested rule, composition of the disadvantaged and comparator groups, and the nature, severity, and significance of the alleged harm. Attempts to automate fairness under EU non-discrimination law similarly must account for this additional complexity in contextual interpretation of the law.

## IV.   CONSISTENT ASSESSMENT PROCEDURES FOR AUTOMATED DISCRIMINATION

Our analysis of jurisprudence of the ECJ and national courts in the EU revealed that the usage of statistical evidence to prove *prima facie* discrimination is rare and inconsistent, lacking well-defined and standardised thresholds for illegal disparity which would hold across different cases. Rather than a consistent top-down approach, fairness has historically been specified contextually according to the details of a case. Fairness is defined by judicial intuition, not statistics. The courts do not provide a consistent and coherent approach to assessing *prima facie* discrimination. As a result, system developers, controllers, regulators, and users lack clear and consistent legal requirements that could be translated into system design and governance mechanisms to detect, remedy, and prevent automated discrimination.

Recognising the higher legal protections in place for special category data,[252] as well as the greater capacity to identify new proxies for traditionally protected attributes brought about by AI,[253] direct discrimination cases are in guaranteeing a minimum replacement income; need to respond to the demand for minor employment and to fight unlawful employment" are legitimate interests.

[250] TOBLER, *supra* note 109 at 41; ELLIS AND WATSON, *supra* note 81; CASE C-1/95, *supra* note 73 at 35.

[251] On legitimate interests for indirect discrimiantion and their limits Wachter, *supra* note 10 at 50–58 for example it is unlikley that purley economic reasons and business considerations (e.g. comsumer satisfaction) alone can justify indirect disrimination.

[252] Wachter and Mittelstadt, *supra* note 31.

[253] Barocas and Selbst, *supra* note 5; O'NEIL, *supra* note 4; Mittelstadt and Floridi, *supra* note 34.

all likelihood going to be rare in automated systems. Rather, the risk introduced by these systems is of widespread and subtle indirect discrimination. As a result, statistical evidence is likely to grow in importance as the potential for automated indirect discrimination grows.[254] The rapid growth in research on statistical measures of fairness among the data science and machine learning community in recent years reflects international consensus that bias and discrimination in automated systems is a growing problem requiring new detection methods and forms of remedy.[255]

A.    TOWARDS CONSISTENT ASSESSMENT PROCEDURES

Assuming statistical evidence will be an essential tool to identify and assess cases of *prima facie* automated discrimination, a natural question to ask is which types of statistical evidence will be necessary or most helpful. As indicated throughout the preceding discussion, this question cannot be answered at a general level. EU non-discrimination law is fundamentally contextual. Judicial interpretation is required to give meaning to key concepts and to answer key questions that affect how discrimination is conceptualised and the evidence needed to support and refute claims. Answers are typically set a national, regional, or case-specific level according to local social, political, and environmental factors. National laws rarely define specific tests, thresholds, or metrics of illegal disparity,[256] or when define variation exists across Member States. 'Contextual equality' as such has, by design, led to a fragmented protection standard across Europe.[257]

---

[254] It could be argued that the growing importance of statistical evidence will shift contextual normative decisions traditionally made by the judiciary (e.g. composition of disadvantaged and comparator groups, evidence of a 'particular disadvantage', disparity thresholds) to the system controllers capable of producing statistics which would demonstrate (or disprove) disparity caused by automated systems. This is a concern that deserves serious consideration: it is very likely that, without regulatory intervention, only system controllers will have the access and expertise necessary to produce the statistical evidence needed to assess potential automated discrimination. This possibility does not, however, undermine the need for consistent procedures for assessment, and related standards for high quality statistical evidence that meets as far as possible the comparative 'gold standard' set by the ECJ (see: Section III.E).

[255] SOLON BAROCAS, MORITZ HARDT & ARVIND NARAYANAN, FAIRNESS AND MACHINE LEARNING, https://fairmlbook.org/ (last visited Feb 28, 2020); Verma and Rubin, *supra* note 8; Friedler et al., *supra* note 8; Kusner et al., *supra* note 8; Pleiss et al., *supra* note 8.

[256] FARKAS AND O'DEMPSEY, *supra* note 7 at 49.

[257] *Id.* at 37.; FARKAS ET AL., *supra* note 7 at 37. Inconsistency across Member States may pose practical challenges but it is not an unintended or negative side effect of the law. Rather, as Directives require Member State implementation and interpretation, fragmented standards are to be expected.

At a minimum, the following normative questions and concepts have historically been interpreted at a national, regional, or local level by the judiciary:

- What is the reach of the contested rule?
- Who are the disadvantaged group(s)?
- Who are legitimate comparator groups?
- Is the nature, severity, and significance of the alleged harm sufficient to be considered a 'particular disadvantage'?
- Which types of evidence are admissible and relevant to assess the case?

Barring future changes in relevant non-discrimination law or new EU regulations, this contextual normative flexibility will need to be respected when building considerations of fairness and non-discrimination into automated systems. To date, much work has been undertaken by the technical community to answer the normative questions above, seen for example in the development of innumerable fairness metrics that implicitly propose thresholds for acceptable and unacceptable disparity.[258] This work can help the judiciary understand and contextualise illegal disparity (see: Section III.C), but it cannot and should not be used to answer these normative and legal questions on a general, case- or application-neutral level. Only the judiciary and regulators possess the democratic legitimacy necessary to justifiably answer such questions.[259]

This is not to suggest that the technical community should not have a voice in debating the right course of action to address such concerns. The technical community has a vital role to play in providing statistical evidence and in developing tools for detection of bias and measuring fairness. But the concept of "contextual equality" needs to be guaranteed and exercised by the judiciary, legislators and regulators. What must be avoided is a situation in which system developers and controllers alone set normative thresholds for discrimination locally and subjectively without external regulatory or judicial input. The case-by-case judgement of the ECJ and national courts should not be replaced wholesale by the system-by-system preferences of developers and system controllers. Such a situation would undermine how non-discrimination law is practiced in Europe.

---

[258] Reuben Binns, *Fairness in machine learning: Lessons from political philosophy*, ARXIV PREPR. ARXIV171203586 (2017); Verma and Rubin, *supra* note 8; Corbett-Davies and Goel, *supra* note 8; Friedler et al., *supra* note 8; Pleiss et al., *supra* note 8. For example, a considerable amount of work is inspired by the 'four-fifths rule' in U.S. law which describes legally acceptable disparity. The rule, which was created as a 'rule of thumb' to assess employment disparity, is generally seen as a positive example that offers clear-cut, automatable decision-making rules. The rule was first published in 1978 in the Uniform Guidelines on Employee Selection Procedures by the U.S. Civil Service Commission, the Department of Labor, the Department of Justice, and the Equal Employment Opportunity Commission. See: Barocas and Selbst, *supra* note 5.

[259] Michel Rosenfeld, *The rule of law and the legitimacy of constitutional democracy*, 74 CAL REV 1307 (2000).

To reconcile this tension, we propose bringing together the strengths of both the technical and legal communities. The need to design systems that accommodate the interpretive flexibility and democratic legitimacy of the judiciary to decide such normative matters does not preclude considerations of fairness and non-discrimination from being included in system design or governance. Rather, the judiciary and legal community needs help from the technical community to develop practical tools and statistical measures that enable contextual equality for automated discrimination.

While application of the law varies according to judicial interpretation of these normative questions and concepts, some procedural consistency can be recognised which can serve as a foundation for this cross-disciplinary collaboration. First, at a minimum statistical evidence is potentially relevant and admissible (see: Section III.D.1). Assuming automated discrimination can only be assessed at a statistical level, admission of statistical evidence will likely grow. Second, the ECJ has defined a 'gold standard' for comparative assessment according to which the composition and effects experienced by both the disadvantaged and advantaged group should be assessed to develop a full picture of the magnitude of disparity (see: Section III.E).

Rather than viewing fairness as a problem to be solved through automation or technical fixes alone, the technical community should embrace the challenge as a starting point for collaborative investigation. Systems cannot and should not be designed to automatically detect, evaluate, and correct for discriminatory decision-making independent of local guidance and interpretation from the judiciary. Rather, what is required is an 'early warning system' for automated discrimination. This can be achieved by designing systems to automatically or consistently produce the types of statistical evidence necessary for the judiciary to make well-informed normative decisions, and for system controllers to systematically detect potential discrimination before it occurs. In other words, what is required are consistent technical standards that align with the judiciary's 'gold standard' procedures for assessing *prima facie* discrimination.

The judiciary has historically relied upon intuitive measures and tests to implement this 'gold standard' and assess *prima facie* discrimination. The law is designed to prevent known forms of human prejudice and discrimination, such as gender inequality in the workplace. Jurisprudence has evolved around real life cases where judges have understandably primary used intuitive measures and tests to deal with known issues. In the past judges did not always need to rely on statistics because the contested rule[260] appeared discriminatory 'on the face'. When dealing with intuitive forms of discrimination, a "flexible and pragmatic" approach to assessment is sensible.[261] However, intuition is poorly suited to detect automated discrimination that does not (necessarily) resemble human discrimination, and can be altogether more subtle, widespread, and based on complex patterns and correlations perceived to exist between people. The judiciary has not yet had the chance to develop tools and consistent procedures that respond to the emergent and novel challenges posed by automated discrimination. This is precisely the accountability gap the technical community can help solve going forward by designing systems and tools that support consistent detection and assessment of *prima facie* automated discrimination.

Coherent procedures to detect and assess *prima facie* automated discrimination are urgently needed to support judges, regulators, industry, and claimants. These tools are essential to ensure effective remedies are available for potential victims.[262] In the remaining sections we analyse how prior work on statistical measures of fairness produced by the technical community measures up against legal procedures for assessing *prima facie* discrimination, and in particular the 'gold standard' for comparing groups proposed by the ECJ.

## V.     STATISTICAL FAIRNESS AND THE LEGAL 'GOLD STANDARD'

The preceding discussion focuses on issues arising from statistical based testing for indirect discrimination under EU law. Flexible, agile, but inconsistent tests for discrimination have emerged across ECJ and Member State jurisprudence.[263] Given the lack of clear legislative guidance and the need to accommodate contextual equality, it is unsurprising that judges have often shied away from requiring statistical evidence to establish *prima facie* discrimination, preferring instead to reason directly about influences using a mixture of common sense and direct reasoning about the impact of contested rules (see: Section III.D). As a result, system developers, controllers, and regulators seeking to evaluate whether a given system or set of outputs is potentially discriminatory, or whether their systems conform to relevant non-discrimination law, lack a consistent set evidential standards that would specify which types of statistical tests should be run and the groups they should compare.[264]

---

[260] This stems from the fact that real-life cases have been brought to the Court of Justice which deal with rather obvious, albeit often indirect, causes of inequality. For example, with favourable treatment of part-time workers it was, historically speaking, very clear that this will affect women more than men. See: CASE C-1/95, *supra* note 73.

[261] TOBLER, *supra* note 109 at 41; MAKKONEN, *supra* note 38 at 36. The judiciary's preference for agile and contextual assessment is sensible when the magnitude of discrimination depends on the type of harm, whether a minority is concerned, or whether discrimination constitutes a systemic injustice that has prevailed over a long period (i.e. the 'facts of the case'). Static requirements for admissible and convincing evidence, or strict high-level thresholds for illegal disparity are not conducive assessing matters of equality that differ substantially in terms of the nature, severity, and significance of harms (see: Section III.C).

[262] Existing complaint procedures are infrequently used across Member States, which may reflect that such procedures are not widely known among potential claimants and representative bodies, or that the possible remedies are not seen as sufficiently meaningful or effective to pursue a case. See: FREDMAN, EUROPEAN COMMISSION, AND EUROPEAN NETWORK OF LEGAL EXPERTS IN THE FIELD OF GENDER EQUALITY, *supra* note 198 at 2.

To automate fairness or detection of *prima facie* discrimination at scale, consistent requirements are seemingly needed. This observation raises a question: is there an approach that could identify potential discrimination in algorithmic systems while still respecting contextual equality as practiced by courts in Europe? This section considers this question in the context of existing work on fairness in automated systems.

A.      STATISTICAL FAIRNESS TESTS IN EU JURISPRUDENCE

Fixed requirements for statistical evidence and thresholds of illegal disparity have not been set in EU jurisprudence.[265] With that said, where statistical evidence has been used to assess whether discrimination has occurred, two general types of tests have been used. These judiciary tests have closely related counterparts in statistical fairness metrics. In this section we examine the practical effects of these tests in terms of the restrictions they impose on how claims of *prima facie* discrimination can be raised, and by whom.

Compared to many of the measures used in algorithmic fairness such as equalised odds[266] or calibration,[267] the tests used in EU jurisprudence and the measure that we advance in the next section are designed to work in an extremely broad set of scenarios. In particular they are designed for cases where there is no "ground-truth data" to compare the decisions made by the system against, meaning tests based on the accuracy of the system with respect to particular protected groups cannot be applied.

---

[263] FARKAS AND O'DEMPSEY, *supra* note 7 at 37; FARKAS ET AL., *supra* note 7 at 37.

[264] Verma and Rubin, *supra* note 8; Pleiss et al., *supra* note 8; Jon Kleinberg & Manish Raghavan, *Selection Problems in the Presence of Implicit Bias*, 94 *in* 9TH INNOVATIONS IN THEORETICAL COMPUTER SCIENCE CONFERENCE (ITCS 2018) 33:1–33:17 (Anna R. Karlin ed., 2018), http://drops.dagstuhl.de/opus/volltexte/2018/8323.

[265] EUROPEAN UNION AGENCY FOR FUNDAMENTAL RIGHTS AND COUNCIL OF EUROPE, *supra* note 141 at 242–243.

[266] Hardt, Moritz, Eric Price, and Nati Srebro. "Equality of opportunity in supervised learning." *Advances in neural information processing systems*. 2016.

[267] Pleiss, G., Raghavan, M., Wu, F., Kleinberg, J., & Weinberger, K. Q. (2017). On fairness and calibration. In Advances in Neural Information Processing Systems (pp. 5680-5689). (should be cited elsewhere)

Following the algorithmic fairness literature, we refer to the first of the two jurisprudence tests as 'Demographic (Dis)parity'. Demographic parity asserts that the proportion of people with a protected attribute should be the same in the advantaged and disadvantaged group, and that consequently, this proportion must match the demographic statistics across the population as a whole. For example, if 35% of white applicants to graduate school are admitted, demographic parity would be satisfied only if 35% of black applicants were also admitted. Systems that do not satisfy demographic parity are said to exhibit demographic disparity.

We refer to the second approach implicitly used by the judiciary,[268] but not yet advanced in the algorithmic fairness literature, as 'negative dominance'. This is a two-part test that asserts that (1) the majority of the disadvantaged group should not be from a protected class, and (2) only a minority of the protected class can be in the advantaged group. In non-discrimination cases the threshold for illegal disparity is typically set higher than a simple majority (e.g. 80 or 90% majority in the disadvantaged group),[269] and the second part of the test would typically assumed to be true.[270] Taking the above example,

---

[268] Specifically, the measure is implied through the comparative language used in the following cases: CASE C-363/12, *supra* note 142 at 53; CASE C-1/95, *supra* note 73 at 30, 33; CASE C-123/10, *supra* note 142 at 56; CASE C-7/12, *supra* note 131 at 39; CASE C-527/13, *supra* note 143 at 28; CASE C-385/11, *supra* note 143 at 29, 31; CASE C-170/84, *supra* note 68; CASE C-256/01, *supra* note 65; CASE 171/88, *supra* note 69; CASE C-703/17 OPINION OF ADVOCATE GENERAL LEGER, *supra* note 98 at 61; CASE C-668/15, *supra* note 115 at 30; CASE C-83/14, *supra* note 121 at 101; CASE C-161/18, *supra* note 71 at 38; JOINED CASES C-399/92, C-409/92, C-425/92, C-34/93, C-50/93 AND C-78/93, *supra* note 74; CASE C-33/89, *supra* note 149 at 89; CASE 171/88, *supra* note 69; CASE C-300/06, *supra* note 73; CASE C-167/97, *supra* note 158; JOINT CASES C-4/02 AND C-5/02, *supra* note 161 at 63–64. See also: Section III.C.

[269] In fact, this form of testing is largely unique to EU law. For historic reasons, EU anti-discrimination legislation's initial primary focus was upon sexual discrimination, while the US anti-discrimination legislation has been strongly tied to problems of racial discrimination. As we show, the shortcomings of negative dominance are much more apparent when dealing with groups of different sizes, as is common in cases of racial discrimination. Recognising this, it is perhaps unsurprising that tests relating to demographic disparity (or disparate impact) are preferred in the US.

[270] The second part of the test can be assumed to be true on the basis that a claim of significant disparity on protected grounds was raised at all. Such a claim would be *prima facie* odd if the protected class in the disadvantaged group also composes a majority of the advantaged group (e.g. 'applicants not admitted' consists of 70% white males but 'applicants admitted' consists of 80% white males). Such cases would typically mean that the alleged

negative dominance would be satisfied as long as black applicants did not make up the majority ($n > 50\%$) of rejected applicants (i.e. the disadvantaged group).

Writing $A$ for the proportion of people belonging to the protected class in the advantaged group, and $D$ for the proportion of people belonging to the protected class in the disadvantaged group, i.e.

$$A = \frac{\text{No. of protected people in advantaged group}}{\text{Total no. of people in advantaged group}}$$

$$D = \frac{\text{No. of protected people in disadvantaged group}}{\text{Total no. of people in disadvantaged group}}$$

We can write the two tests as the following formulas:

| | |
|---|---|
| **Demographic disparity**[271] | $D > A$ |
| **Negative dominance** | $D > 50\% > A$ |

Looking at the two formulas it is immediately obvious that negative dominance is a much harder test to satisfy. As a result, whenever a test shows that negative dominance exists, demographic disparity by definition is also present.

The key insight into understanding these two tests and the interplay between them is two-fold:

1. If the group with a protected attribute occupies fifty percent of the population the two tests are functionally the same.

---

disadvantaged protected group made up the majority of the affected population overall (e.g. applicants). Again, this is a 'rule of thumb' assumption rather than a universal truth for cases in which negative dominance is used to assess potential discrimination.

[271] In some situations it may be challenging to directly measure one of either $D$ or $A$, but the proportion of people belonging to the protected group in the affected population (i.e. the population of people that could either be advantaged or disadvantaged by the contested rule) is known. Writing $P$ for this proportion demographic disparity can also be identified if either $P > A$ or $D > P$.

2.   In the early sexual discrimination cases that set precedence,[272] approximately fifty percent of the population were women and members of the group with a protected attribute.

As such, many precedent establishing cases made use of the negative dominance. However, we argue that negative dominance is an inappropriate test in scenarios involving a minority group, or one where members of the protected group make up less than fifty percent of the population.

To unpack the implications of the two tests when applied to cases involving minority groups, we consider two hypothetical companies:

- Company A is composed of 55 women and 55 men.
- Company B is composed of 100 hundred white people and 10 black people.

For this thought experiment, we will assume that each company has made a decision that disadvantages the entirety of the relevant protected subgroup (i.e. all 55 women in company A and all 10 black people in company B). We will then ask how many men or white people (respectively) each company would need to disadvantage under the two tests for the disparity to be considered non-discriminatory without the company offering any further justification.

For Company A, this is a straightforward task. As 100% of women lie in the disadvantaged group, demographic parity will only occur when 100% of men are also in the disadvantaged group. Similarly, negative dominance also requires at least 50% of the negative group to be composed of men, which can only happen if 100% of men are disadvantaged. In this case because the company is composed of two groups of equal size, roughly matching the gender distribution in the general population, the two tests impose identical requirements.

In contrast, for Company B the two tests lead to substantially different requirements for the disparity between protected groups to be considered non-discriminatory. As before demographic disparity only occurs when 100% of white employees are disadvantaged, matching the ratio of disadvantaged black people. In contrast, negative dominance requires that only a matching number of white people are disadvantaged. In this case the disparity would be non-discriminatory if at least 10% of the white people in Company B were similarly disadvantaged.

As an additional example of negative dominance failing to reveal implicit discrimination, we consider a hypothetical example of implicit discrimination inspired by UK case law, namely that refusing to hire men with beards is a form of religious discrimination against Sikhs.[273] As of 2011, 37% of men in the UK had a form of facial hair,[274] while 0.8% of the entire population in England and Wales were practicing Sikhs.[275] Assuming 0.8% of the male adult population are Sikh and using the definitions above, we have $n = \frac{0.8\%}{37\%} = 1.9\%$ and $p = 0\%$. These ratios clearly satisfy the definition of demographic disparity, while falling far short of the threshold of $n > 50\%$ required for negative dominance.

---

[272] See: Footnote 268.

B.        SHORTCOMINGS OF NEGATIVE DOMINANCE

Statistical testing cannot fully replace the intuition used by the courts. However, if consistent application of non-discrimination across Member States is sought, it is nonetheless vital that in situations where statistical measures or intuition could be used, the statistical measures must be chosen to be consistent with our intuition as to what types of disparity count as unfair, or discriminatory.

As shown by the above examples, the fewer people who have a particular protected attribute, the more pronounced problems with negative dominance become. This has the counter-intuitive effect of the protection offered to groups under non-discrimination law being dependent on the size of the group. In other words, minorities receive less protection precisely because they are minorities. This makes the test for negative dominance particularly ill-suited for issues of ethnicity and sexual orientation.

One particularly worrying issue is that the use of negative dominance opens the door to a possible novel justification of disparity we call 'divide and conquer'. With this approach it would be possible to defend the behaviour of a system that exhibits negative dominance against a group by splitting the disadvantaged group into subgroups and showing that each subgroup is so small that none satisfy negative dominance. For example, by finding a justification to split a group of women into intersectional groups of black women, white women, Asian women, and so on, and then running statistical tests with these subgroups, a company could argue a contested rule does not satisfy negative dominance and thereby justify disparity even if the test taken in aggregate over all women would show negative dominance.

---

[273] Examples taken from UK case law see MAKKONEN, *supra* note 38 at 34; ELLIS AND WATSON, *supra* note 81 at 168.

[274] YouGov, *Beards are growing on the British public* (2017), https://yougov.co.uk/topics/politics/articles-reports/2017/03/10/beards-are-growing-british-public (last visited Feb 22, 2020).

[275] Office for National Statistics, *Religion in England and Wales 2011* (2011), https://www.ons.gov.uk/peoplepopulationandcommunity/culturalidentity/religion/articles/religioninenglandandwales2011/2012-12-11 (last visited Feb 22, 2020).

In comparison, demographic disparity has none of these issues as it simply occurs if $k$% of the population are in the advantaged group, and more than $k$% are in the disadvantaged group. As such, use of intersectionality under testing for demographic disparity would, in contrast to negative dominance increase the burden on the alleged offender. Demographic disparity is thus a preferable measure as it does not create a loophole to justify potentially discriminatory rules which circumvents the intention of existing legislation. Negative dominance allows for precisely the sort of "battle of numbers" the court has worried about in the past.[276]

Given the inconsistency in testing methods and thresholds applied by the judiciary, what if anything should be done to resolve the legal ambiguity? We argue that given a choice between the two tests of demographic disparity and negative dominance, the judiciary should unambiguously select demographic disparity as its accepted standard for statistical evidence. Demographic parity facilitates a realistic comparison between groups which aligns with the spirit of the 'gold standard' set by the ECJ. Setting such a requirement is already within the judiciary's power. The non-discrimination directives were written with sufficient generality to support many forms of testing, and while early cases made use of negative dominance for inferring matters of sexual discrimination, such precedence should not be binding on future cases.

## VI.  CONDITIONAL DEMOGRAPHIC DISPARITY: A STATISTICAL 'GOLD STANDARD' FOR AUTOMATED FAIRNESS

Based on our analysis of ECJ and Member State non-discrimination jurisprudence, and taking into account the varied and contextual evidential standards for *prima facie* discrimination that have emerged over time, we propose 'conditional demographic disparity'[277] as a minimal standard for statistical evidence in non-discrimination cases addressing automated systems. Much work on fairness in the machine learning community has been dedicated to defining statistical metrics for fairness which can be applied to the outputs of automated systems to detect *prima facie* discrimination. In many cases, this work is seemingly headed towards a point at which discrimination can be detected and corrected entirely within the system development lifecycle.

---

[276] TOBLER, *supra* note 109 at 43; CASE C-137/93 OPINION OF ADVOCATE GENERAL LEGER, *supra* note 163 at 53.

[277] Faisal Kamiran, Indrė Žliobaitė & Toon Calders, *Quantifying explainable discrimination and removing illegal discrimination in automated decision making*, 35 KNOWL. INF. SYST. 613–644 (2013).

Such an approach would not, however, align well with the notion of 'contextual equality' in non-discrimination law in Europe in which judicial interpretation plays a significant role. Recognising this, we encourage the technical community to embrace the flexible nature of equality and fairness in European law, recognise its strengths in terms of responsiveness to local social and political factors of non-discrimination cases, and find ways to translate the approach into complementary technical tools and organisational measures. In particular, we hope that the technical community will dedicate further effort to designing mechanisms capable of producing standard types of statistical evidence that can help the judiciary to flexibly assess potential automated discrimination. These mechanisms would also be highly valuable for system controllers and developers to conduct internal pre-emptive testing for bias and disparity in their systems.

Re-aligned in this way, the proper goal of 'automating fairness' would not just be to correct potential discrimination as a part of system development and deployment, but rather to create systems which can automatically provide the baseline evidence required for external adjudication. This re-alignment will help the judiciary continue to fulfil its crucial role as an interpreter of non-discrimination law on a contextual, case-specific basis and help system controllers and developers to pre-emptively correct biases and potential discriminatory results.

We propose 'conditional demographic disparity' as a standard for statistical evidence that is harmonious with the aims of EU non-discrimination legislation as well as the 'gold standard' for statistical evidence previously set by the ECJ. If adopted as an evidential standard, CDD will help answer the two key questions concerning fairness in automated systems that can be justifiably delegated to the machine learning community. Specifically, in any given case of *prima facie* discrimination caused by an automated system:

1.  Across the entire affected population, which protected groups could I compare to identify potential discrimination?
2.  How do these protected groups compare to one another in terms of disparity of outcomes?

CDD answers both of these questions by providing measurements for making comparisons across protected groups in terms of the distribution of outcomes. This information is essential for the judiciary to assess cases of potential automated discrimination in several senses. First, it can help identify potentially illegal disparity between specific groups in a population requiring further examination and justification. Inter-group comparisons can help the judiciary define the composition of both the disadvantaged and comparator groups which must account for normative, political, and social aspects of the case. Second, it enables exercise of the Court's 'gold standard' for assessment of *prima facie* discrimination by providing sufficient information to examine both the disadvantaged and advantaged groups.

Conditional demographic (dis)parity was first proposed as a measure by Kamiran et al.[278] Writing $A$ for the proportion of people belonging to this protected class in the advantaged group with attributes $R$, and $D$ for the proportion of people belonging to the protected class in the disadvantaged group with attributes $R$, i.e.

$$A_R = \frac{\text{No. of protected people in advantaged group with attributes R}}{\text{total no. of people in advantaged group with attributes R}}$$

$$D_R = \frac{\text{No. of protected people in disadvantaged group with attributes R}}{\text{Total no. of people in disadvantaged group with attributes R}}$$

We can write the set of tests as the following formula:

| Conditional Demographic Disparity | $D_R > A_R$ for any choice of attributes $R$ |
|---|---|

Conditional demographic disparity differs from demographic disparity only in the sense that one or more additional conditions are added. For example, if we condition on grade point average (GPA) and we admit 70% of white applicants to graduate school that had a 4.0 GPA, conditional demographic parity would be satisfied only if 70% of black applicants with a 4.0 GPA were also admitted.

---

[278] *Id.* CDD is referred to as conditional (non-)discrimination in their paper, however to improve clarity, we use "conditional demographic disparity" given the plethora of fairness definitions in the literature. In the context of this Article "conditional demographic parity" is used to refer to the original test proposed by Kamiran et al. In contrast, we use "conditional demographic disparity" to refer to the measure we propose which does not provide a binary pass/fail answer as a test would, but rather reports on the magnitude of disparity between groups in an affected population. The measure is intended to assist in contextual analysis of potential automated discrimination rather than confirming that illegal disparity has occurred.

Formally, a system satisfies conditional demographic disparity only if there is no choice of attributes for which conditional demographic disparity holds. However, searching over all combinations of attributes for a violation in practice would be prone to finding false positives. As a result, an unbiased system would often appear by chance to be biased in one or several attribute groups. A discussion of statistical tests and aggregate statistics that take this into account is given in Section A.

A reasonable question would be why does this measure satisfy the 'gold standard' established by the Court? After all, a contested automated process potentially depends upon many factors, not just those that statistical testing has been conditioned on, and may be fully automated or have humans in the loop.

To answer this question, we consider a related scenario in which we split the population into groups that share a set of common attributes $R$, and split these groups into subgroups by conditioning on protected attributes, such as ethnicity or sex. If we now rank each of these subgroups according to some appropriate criteria and for each subgroup, selecting the top $k_R\%$ proportion of each subgroup,[279] then this method clearly satisfies the legal requirements. Each subgroup is treated equally within the group, and the proportion of people selected from each subgroup depends only on $R$ as required.[280]

Finally, we observe that any method of assigning people to advantaged and disadvantaged groups that satisfies Conditional Demographic Parity (i.e. $D_R = A_R$ for all $R$) can be interpreted as exactly following the procedure described in the paragraph above. A mathematical proof of this follows from a straightforward sequence of elementary operations reported in Appendix 1.

A.    CONDITIONAL DEMOGRAPHIC PARITY IN PRACTICE

The issues faced in assessing discrimination in a legal context mirror those faced by the statistical community in assessing bias for more than fifty years. Similarly, conditional demographic parity mirrors the solution proposed by the statistical community for understanding the source of bias. To demonstrate the practical contribution of conditional demographic parity we re-visit a study of gender-bias in graduate admissions reported by Bickel et al,[281] which is one of the most famous examples of Simpson's paradox in the statistical literature.

---

[279] Note that $k_R\%$ may vary with the choice of group $R$.

[280] See Kleinberg and Raghavan, *supra* note 285 for a discussion of selection strategies to mitigate bias.

[281] P. J. Bickel, E. A. Hammel & J. W. O'Connell, *Sex Bias in Graduate Admissions: Data from Berkeley*, 187 SCIENCE 398–404 (1975).

The authors were interested in apparent gender-bias in graduate admissions at the University of California, Berkeley.[282] The application pool at Berkeley was already biased in favour of men, with 8442 male applicants to 4321 female applicants. Of these, 44% of the male applicants were accepted compared to only 35% of female applicants. The authors noted that due to the large number of applicants, it was incredibly unlikely that this discrepancy was due to chance alone. At this point, the authors checked whether a particular department was responsible but found that no department exhibited a significant bias against women; instead the issue was that women were much more likely to apply to more competitive departments (such as English) that were much more likely to reject graduates of any gender, whereas other departments (such as Engineering) were more lenient.

To rephrase these conclusions in the language of demographic parity: although Berkeley's pattern of admission exhibited strong evidence of demographic disparity, once we condition according to 'department applied for', the apparent bias disappears. In Table 1 we revisit this example using a subset of the data used by the original paper consisting of redacted publicly available statistics,[283] with department names removed by the university.

| | Admitted | | | Rejected | | |
|---|---|---|---|---|---|---|
| **Department** | *Male* | *Female* | *Total* | *Male* | *Female* | *Total* |
| **A** | 512 | 89 | 601 | 313 | 19 | 332 |
| **B** | 313 | 17 | 330 | 207 | 8 | 215 |
| **C** | 120 | 202 | 322 | 205 | 391 | 596 |
| **D** | 138 | 131 | 269 | 279 | 244 | 523 |
| **E** | 53 | 94 | 147 | 138 | 299 | 437 |
| **F** | 22 | 24 | 46 | 351 | 317 | 668 |
| **Total** | **1158** | **557** | **1715** | **1493** | **1278** | **2771** |

*Table 1 – Berkeley admissions data by department and gender*

Computing the proportion of men and women in the advantaged and disadvantaged groups we arrive at Table 2:

| | Admitted | | Rejected | |
|---|---|---|---|---|
| **Department** | *Male* | *Female* | *Male* | *Female* |
| **A** | 85% | 15% | 94% | 6% |
| **B** | 95% | 5% | 96% | 4% |
| **C** | 37% | 63% | 34% | 66% |
| **D** | 51% | 49% | 53% | 47% |
| **E** | 36% | 64% | 32% | 68% |
| **F** | 48% | 52% | 53% | 47% |
| **Total** | **68%** | **32%** | **54%** | **46%** |
| *Table 2 – Admissions and rejections by gender* | | | | |

As indicated by the 'Total' row which describes behaviour across the university, we find that while men (8442 applicants) applied to Berkeley at almost double the rate of women (4321 applicants), the proportion of women that were rejected by Berkeley (or, are members of the disadvantaged group) is only 46% and noticeably below the 50% negative dominance threshold (see: Section V.A). As such, although this paper is one of the most famous examples of Simpsons's paradox and is taught world over as a clear example of systemic gender bias, it falls short of satisfying the legal test of negative dominance. In contrast, the proportion of women that make up the advantaged group is 32% (which is substantially smaller than the rejected 46%), presenting clear evidence of demographic disparity. This is unsurprising as demographic disparity is a stronger test than negative dominance, meaning it will classify more cases as *prima facie* discriminatory (see: Section V.A).

However, even though this data shows clear evidence of systematic bias in terms of demographic disparity, a justification can be given in terms of conditional demographic parity. As we condition upon the department applied for (rows A-F) the statistics become considerably less clearly discriminatory. Instead, we find fluctuations in the proportion of women in the admitted and rejected groups, with women sometimes making up a greater proportion of the rejected group than the admitted, and sometimes vice versa. The largest fluctuations occur in department A where few women applied. Such fluctuations are expected by chance and cannot be taken as *prima facie* evidence that a particular department is biased either in favour of or against women.

To account for these fluctuations and explain their statistical power, Bickel et al. computed aggregate statistics over all departments using various approaches including a method proposed by Fisher.[284] These aggregate statistics revealed substantial evidence of an asymmetric bias across departments. However, the bias revealed was against men, not women.[285]

Freedman et al. adopted a simpler approach that used the weighted average of the departmental statistics.[286] They weighted each departmental statistic by the number of total applicants to the department. This gives rise to the following final summary statistic, which is conditional on the department applied to by applicants (see: Table 3).

| Conditionally Admitted | | Conditionally Rejected | |
|:---:|:---:|:---:|:---:|
| *Male* | *Female* | *Male* | *Female* |
| 58% | 42% | 60% | 40% |
| *Table 3 – Admissions data conditioned on department* | | | |

This result is consistent with the analyses conducted by Bickel et al. and Freedman et al. which show a small bias in favour of women. In other words, after conditioning on the department applied for, women make up a slightly larger proportion of the advantaged groups than the disadvantaged groups across departments and the university as a whole.

This well-known and paradoxical result leads naturally to the question as to how much bias is acceptable, and where a threshold should be set for illegal disparity, or a "particular disadvantage." As discussed in Section III.C, setting such a threshold under EU non-discrimination law is a fundamentally contextual and political question traditionally answered by national courts and the ECJ. In terms of designing unbiased or non-discriminatory AI systems, no system can be perfectly calibrated and future-proof. For systems working on real problems, or using real-world data, at a minimum some small amount of bias in one direction or another should be expected. As we move to the era of 'Big Data' and large multi-national companies and government agencies deploying systems that affect millions of people, any such bias can quickly become statistically significant. However, an effect of extremely small power can become statistically significant simply by virtue of the large number of people measured, and the size of the effect may remain so small as to be meaningless with regard to any one individual's life. Medical professionals face a similar challenge in bridging the gap between clinically significant and statistically significant treatment effects, with the former being a much higher bar than the latter.[287]

---

[284] Ronald Aylmer Fisher, *Statistical methods for research workers*, *in* BREAKTHROUGHS IN STATISTICS 66–70 (1992).

[285] Bickel, Hammel, and O'Connell, *supra* note 303.

[286] FREEDMAN, PISANI, AND PURVES, *supra* note 305 considered the weighted average of the per department acceptance rate for men and women. For consistency with the legal and fairness literature, we keep the four numbers.

One final question is whether the tests for conditional demographic parity must always be performed over all sets of attributes, and aggregated as discussed, or if they can ever be restricted to a single choice of attribute. According to best practices in statistics, the answer is again contextual. Where, prior to looking at statistical information, convincing information exists that a system is likely to exhibit a bias towards a protected group, this group should be tested for bias in isolation. Doing so prevents noise from the groups of people associated with other attributes overwhelming evidence of bias. However, isolated testing must not performed over all possible groups, as by pure chance data regarding some of the groups is likely to exhibit a non-persistent bias.[288] In situations where no information exists to prefer testing of one group over another, aggregate testing similar to the methods discussed in this section should be preferred.

In deciding when a bias or resulting disparity is too significant and thus illegal, the courts will need to define thresholds, reasonable factors to condition on in CDD testing, and the 'reach' of contested rules or systems (see: Section III.B). Each of these decisions needs to respect the autonomy of decision-making entities while sufficiently protecting individuals and groups against the harm caused by the system. Just as the decision as to which factors it is reasonable to condition on is context-dependent, the decision as to how much disparity is too much is also highly context-dependent according to the purpose of the decision-making system, the harms associated with it, and the number of people affected. As such, we advocate for the use of summary statistics as discussed by Freedman et al. (see: Table 4) over the tests used by Bickel et al. which only reveal the statistical significance of the bias without giving any indication of its magnitude.

---

[287] Alan E. Kazdin, *The meanings and measurement of clinical significance.* (1999).

[288] Juliet Popper Shaffer, *Multiple hypothesis testing*, 46 ANNU. REV. PSYCHOL. 561–584 (1995). For a compelling visualisation of these issues see: XKCD, *Significant*, XKCD (2020), https://xkcd.com/882/ (last visited Feb 24, 2020).

B.    SUMMARY STATISTICS TO SUPPORT JUDICIAL INTERPRETATION

The preceding sections have analysed how to connect legal definitions of equality in EU non-discrimination law with existing methods from statistics designed to measure unfairness despite confounding influences. Based on our analysis, we recommend CDD as a baseline evidential standard to produce summary statistics for assessing potential automated discrimination. While the measure we argue for is referred to as conditional demographic parity in the field of algorithmic fairness,[289] the measure itself predates algorithmic fairness as a field, and has been used for more than 45 years as a way to understand the effects of bias in the face of socially acceptable confounders.[290]

The departure we make from these prior approaches is to advocate for a use of descriptive summary statistics to allow a semi-technical audience to understand the magnitude of the effect, rather than statistical tests. These descriptive measures are themselves standard and wide-spread, and used in established textbooks for summarising the biases we are interested in modelling.[291] Our reason for advocating for these descriptive measures over statistical testing is to avoid the "battle of numbers" that have worried legal experts,[292] and instead to clearly describe the magnitude of the bias. This form of reporting will allow the judiciary and regulators to use these tools to contextually judge whether disparity in a given case is legally acceptable.

The preceding example of CDD in practice reveals the shortcomings of both negative dominance and non-conditional demographic disparity. Despite being one of the most well-studied cases of gender discrimination, these well-established tests fail to fully capture the nature, severity, and significance of the disparity. This finding reveals how these statistical tests used in prior jurisprudence can offer an incomplete view of statistical disparity, but also how the significant latitude offered by the law to potential offenders could create ethically significant disparity between protected groups without being flagged up as potentially discriminatory.

The utility of CDD should not be overstated. CDD cannot assess whether illegal disparity has occurred or whether illegal disparity is justified; only the judiciary can assess such matters. Rather, summary statistics reporting on CDD provide a roadmap for further contextual investigation by showing the relationship between groups in an affected population and how their size and outcomes compare. Unlike the statistical tests discussed earlier (see: Section V.A), CDD as a statistical measure (as opposed to a test) does not produce a binary yes/no answer as to whether a particular outcome is discriminatory. Rather, CDD will assist in contextual evaluations of disparity that can account for the apparent magnitude of bias and weigh it against competing rights, evidence, or justifications. Used in this way, CDD can help the judiciary, regulators, system controllers, and individuals identify disadvantaged groups, comparator groups, and 'particular disadvantages' (see: Section III) across the entire population affected by a system.

---

[289] Kamiran, Žliobaitė, and Calders, *supra* note 299.
[290] Bickel, Hammel, and O'Connell, *supra* note 303.
[291] FREEDMAN, PISANI, AND PURVES, *supra* note 305.
[292] TOBLER, *supra* note 109 at 43; CASE C-137/93 OPINION OF ADVOCATE GENERAL LEGER, *supra* note 163 at 53.

Adoption of CDD as a standard intended to assist rather than replace or automate judicial decision-making can help close these gaps and better align work on fairness in machine learning with EU non-discrimination law. At the same time, CDD might also help inform the reasoning of the legal community in cases where discriminators are algorithms. If adopted as a baseline evidential standard, CDD can ensure that both claimants and alleged offenders can refer to a common set of statistical evidence. This would help reduce the burden on both parties to produce relevant and convincing statistics, which is often beyond the resources of individual claimants.[293] Cases of intuitive discrimination are likely to decrease in cases of automated decision-making, meaning without novel tools to detect bias and disparity, serious instances of potential discrimination may be missed.

Summary statistics reporting on conditional demographic disparity can provide the baseline evidence required to detect this sort of heterogenous, minority-based and intersectional discrimination which might otherwise be missed through the judiciary's current approach to assessing potential discrimination. In proposing CDD as a common 'gold standard' for the technical and legal communities to advance work on automating fairness we are not, however, advocating for any particular answer to the fundamentally normative determinations (i.e. legality thresholds, conditioning factors, and the 'reach' of the rule) that influence the scope and application of such statistics. Such determinations can only justifiably be made at a case-specific, local level (see: Section III). Our aim is to avoid a situation in which AI systems, or their developers and controllers, set thresholds of acceptable and illegal disparity by default. Such a situation could be problematic due to a lack of democratic legitimacy. We propose CDD as a means to provide the judiciary, regulators, system controllers and developers, and claimants with essential baseline information necessary to ensure the magnitude and context of automated discrimination can be fully appreciated.

---

[293] Barnard and Hepple, *supra* note 137; see also Browne, *supra* note 180.

## VII.    AUTOMATED FAIRNESS: AN IMPOSSIBLE TASK?

AI bias is posing a significant challenge in modern society. Computer scientists are struggling to find suitable fairness metrics that are legally compliant and yet static enough to be encoded. This quest is difficult because the law and jurisprudence conceptualise fairness and discrimination as fundamentally contextual concepts. At the same time, increasing use of algorithms is disrupting traditional procedures and remedies against the prevention and investigation of discrimination cases. This paper is an attempt to show that the legal and technical communities could benefit from engaging in a closer dialogue about how to design European considerations of equality and fairness into AI and governance. The legal community could benefit from some of the coherent and static strategies put forward by the technical community to measure unintuitive disparity and bias, while the technical community will need to account for the contextual and flexible nature of equality in Europe when designing systems and governance mechanisms.

In this paper we have argued three points: (1) fairness is contextual and cannot (and arguably should not) be automated in a manner that fully respects 'contextual equality'; (2) AI as a discriminator is disrupting known forms of bias detection, investigation, and prevention and thus traditional intuitive (court) procedures need to be more coherent, static, and explicit in their definitions; and (3) summary statistics reporting on conditional demographic disparity are a first step to close the accountability gap between AI and EU non-discrimination law. We advocate for the usage of CDD as a baseline statistical measure for cases of automated discrimination in Europe. Summary statistics produced using CDD will be of value to judges, regulators, system controllers and developers, and claimants Judges will have a first frame to investigate *prima facie* discrimination, regulators will have a baseline for investigating potential discrimination cases, controllers and developers can prevent discrimination with pre-emptive audits (e.g. when users do not immediately experience and report discrimination) or refute potential claims, and claimants will have a coherent strategy and evidential standard to raise claims in court.

The contextual nature of fairness emerged organically from both case law and legislation. Jurisprudence has evolved around real life cases where judges often use intuitive measures and tests to deal with known issues. Many of those judgements (e.g., *prima facie* discrimination, justification of discrimination) can be decided on logic and common-sense reasoning. For example, to establish *prima facie* discrimination, evidence of a particular disadvantage needs to be provided (see: Section III.C). This is a deeply contextual task that, by design, lacks clear procedural rules, evidential requirements, and comparative definitions (e.g. who is a legitimate comparator). Key definitions, evidential standards, and comparative thresholds are defined against the circumstances of the case, including (1) who is an appropriate disadvantaged and comparator group; (2) what constitutes a particular disadvantage (e.g. financial, missed opportunities), (3) the significance of the discriminatory harm (e.g. must affect a "far greater number" of people, must not be fortuitous, less stricter rules if constant disparity over a long period occurred), (4) whether a concrete harm needs to be proven or a potential threat is sufficient, and (5) which types of evidence are admissible in court (e.g. statistics or common sense assessments).

In general, national and European courts, as well as legal scholars, are reluctant to rely heavily on statistical evidence to prove discrimination claims. Statistics are often not available or not seen as trustworthy. Intuitive or 'on-the-face' evidence, such as "common knowledge,"[294] "obvious facts," [295]or "convictions,"[296] is often preferred. The admissibility of evidence is determined on a case-by-case basis in line with national laws and judicial interpretation, which has created fragmented evidential standards across Europe to assess potential discrimination. Individual national and EU courts have themselves been internally inconsistent over time.[297] Such an ambiguous and inconsistent standard is not fit for automation at scale.

Non-discrimination law is also based on the idea of comparison (see: Sections III.E), another deeply contextual concept. A disadvantaged group or person is treated less favourably because of a protected characteristic in comparison to a group or person receiving preferential treatment. Who is seen as affected by the contested rule (e.g. all job applicants, only qualified job applicants or everyone that saw the job advertisement) and who is considered a legitimate comparator (e.g. marriage versus civil partnership, part-time versus full-time work) are case-specific determinations influenced by social and political factors. Additive and intersectional discrimination further complicate these decisions (see: Section III.A.1). Finally, these determinations must be made with respect to the contradictory preferences, arguments, and evidence offered by the alleged offender in their attempts to refute *prima facie* discrimination (see: Section III.F). Comparative determinations must therefore be justified with respect to the arguments offered by both sides. This need for 'contextual equality' in judicial interpretation further complicates the automation of fairness.

---

[294] ELLIS AND WATSON, *supra* note 81 at 151 which discusses "potentially discriminatory" legislation; CASE C-322/98, *supra* note 202 at 24 noting "it is common ground in Germany."

[295] CASE C-79/99, *supra* note 203.

[296] CASE C-237/94, *supra* note 204.

[297] FARKAS AND O'DEMPSEY, *supra* note 7 at 37; FARKAS ET AL., *supra* note 7 at 37.

In most cases it will be national courts that decide which statistics are admissible, if a disadvantage occurred, if it was significant, and if it can be justified. National courts decide as well whether the "statistical evidence is valid, representative and significant."[298] The fragmented standard across Europe exacerbates the challenge of automating fairness, opening it both to inconsistent detection but also gaming by malicious actors. The inconsistency in standards of admissible evidence makes automating fairness difficult. For example, in running internal audits, which statistical tests and fairness metrics should system controllers give the most weight? Detection of discrimination can be potentially 'gamed' through choice of fairness metrics, but controllers face a double bind: should any and all types of potential discrimination be reported to regulators and users? Is it ethically justifiable to ignore certain statistical tests which cast an automated system in a poor light? How should these determinations be made consistently across the myriad sectors using AI?

Fairness cannot and arguably should not be automated. AI does not have a common sense understanding of contextual equality, cannot capture and consider local political, social and environmental factors of a case, and thus cannot make the type of normative assessments traditionally reserved for the judiciary. This will make it very hard to automate detection and prevention of discrimination.

To bridge this gap the technical community often focuses, among other things, on discussion and development of new fairness metrics. While this rich literature is instrumental to help understand the challenges of AI fairness and bias, these metrics might not yield desired results because they may ultimately prove to be irrelevant in court. To further support and not frustrate the efforts of the technical community, we hope that their focus will expand to include legal guidance provided by jurisprudence. Aligning the design of autonomous systems with contextual equality, flexibility, and the judicial interpretation of the comparative aspects of non-discrimination law would greatly benefit all parties involved.

Even with this shift in focus, further challenges remain for the legal and technical communities. Non-discrimination law is designed to prevent familiar forms of human prejudice and discrimination, such as gender inequality in the workplace or racist hiring practices. The judiciary often relies on intuitive tests to assess discrimination claims. Algorithmic discrimination does not need to follow these familiar patterns of discrimination. Nor does it need to differentiate people according to human perceptible traits or legally protected characteristics. New forms of algorithmic discrimination are increasingly subtle and difficult to detect and may have no basis in the law or jurisprudence, meaning the judiciary has not yet developed proven methods to detect and assess these new types of discrimination. Algorithmic discrimination can be counterintuitive in a way unlike human discrimination which creates a crisis for judiciaries reliant on intuitive assessments. A certain type of proxy might not ring "alarm bells" like other data would have. We know that income can act as proxy for gender, for example, but how do browsing behaviour, likes on Facebook, or clicks on webpages correlate with protected attributes? Judges, regulators and industry might not have an intuitive feeling with these type of data in the same way they had with known proxies such as part-time work. Faced with AI, we can no longer rely on intuition to tell us when and how discrimination occurs.

---

[298] CASE C-161/18, *supra* note 71 at 45.

Potential victims face similar challenges. Proving algorithmic discrimination may be particularly difficult because it will not necessarily be experienced or 'felt' in a manner comparable to human discrimination. Claims can only be raised if a victim actually feels disadvantaged. For example, an applicant may never know when or why their CV was filtered out when applying for a job. A consumer may never know they are not being shown certain advertisements or are receiving comparatively unfavourable product offers. Humans discriminate due to negative attitudes (e.g. stereotypes, prejudice) and unintentional biases (e.g. organisational practices or internalised stereotypes) which can act as a signal to victims that discrimination has occurred.[299] Equivalent mechanisms and agency do not exist in algorithmic systems. Only periodic and preemptive testing will guarantee we are not blind to automated discrimination at scale.

As a result, proven methods for conceptualising and assessing discrimination, which have historically developed through judicial application and interpretation of the law, require an overhaul to ensure algorithmic discrimination does not go undetected, both in individual cases and at a systemic level. Even if automated fairness was technically feasible on a large-scale, eliminating the contextual aspects of fairness would be detrimental to the goal and purpose of non-discrimination law.

---

[299] Makkonen, *supra* note 31 at 57, 64; Cobbe and Singh, *supra* note 31.

    We can only overcome this challenge together. Going forward, judges, regulators and the private sector will increasingly need to work together and adopt systematic, consistent evidential standards and assessment procedures to ensure algorithmic disparity does not go unchecked. Adopting this strategy will help ensure new types of algorithmic discrimination are consistently detected, while still maintaining contextual, case-specific judicial interpretation that is an essential feature of EU non-discrimination law. These standards must serve the needs of the judiciary, regulators, system controllers and developers, as well as individuals and groups affected by automated systems.

    The comparative element of non-discrimination law is most in need of reform. The ECJ has defined a 'gold standard' for comparison for itself, but not consistently used it across prior cases.[300] According to this standard the composition of both the advantaged and disadvantaged groups should always be compared. However, in prior cases the judiciary has often looked only at the composition of the disadvantaged group to assess whether discrimination has occurred.[301] This may be explained by prior trends in demographics and the use of intuitive tests by the judiciary.[302] In the past it was safe to assume, for example, that the statistical distribution of the general population would be reflected inside a company,[303] but this is no longer the case with changes in traditional social and gender roles.[304]  As discussed in Sections III.E and VI, looking only at the disadvantaged group can give an incomplete and misleading picture of the magnitude and relative impact of disparity. For example,

---

[300] According to the Court, "the best approach to the comparison of statistics is to consider, on the one hand, the respective proportions of men in the workforce able to satisfy the requirement of two years' employment under the disputed rule and of those unable to do so, and, on the other, to compare those proportions as regards women in the workforce." CASE C-167/97, *supra* note 71 at 59; CASE C-300/06, *supra* note 73 at 41; TOBLER, *supra* note 109 at 41.

[301] For example, in CASE C-668/15, *supra* note 115; CASE C-170/84, *supra* note 68. In *Maniero*, *Jyske Finans* and *Bilka* the Court did not assess the make-up of both groups in relation to each other.

[302] It is also worth noting that the judiciary's choice to look only at the disadvantaged group in certain cases did not necessarily have detrimental effects, although it has been criticised in this regard.

[303] CASE C-167/97, *supra* note 71; See also UK case law examples in ELLIS AND WATSON, *supra* note 81; See also CASE C-300/06, *supra* note 73; CASE C-1/95, *supra* note 73 regarding the community as a whole.

[304] For example, less favourable rules around part-time work do not necessarily only affect women as has traditionally been the case in Germany. However, in the future it might be the case that the composition of the company does not reflect the general population (e.g. female only start-ups). It might also be the case that even though part-time work is usually carried out by women in Germany, in certain regions like Berlin part-time jobs are mostly taken up by men. See: CASE C-300/06, *supra* note 73.

discrimination would only be revealed by looking at both groups in a case where the disadvantaged group has an equal distribution of members according to ethnicity, but the advantaged group is predominantly white.

Current tests used by the judiciary could incentivise what we call 'divide and conquer' approach. With this approach it would be possible to defend the discriminatory behaviour of a system by splitting the disadvantaged group into subgroups and showing that each subgroup is so small that none of them pass the legally required threshold of significant and disproportionate disadvantage (i.e. negative dominance; see: Section V.B). This would be especially detrimental for cases involving intersectional discrimination or minority groups. The dual group 'gold standard' is thus highly sensible.

This legal gold standard has an equivalent in computer science: conditional demographic (dis)parity (CDD). We propose summary statistics based on CDD as the cornerstone of a coherent strategy to ensure procedural regularity in the identification and assessment of potential discrimination caused by AI and automated decision-making systems. The measure respects contextual equality in EU non-discrimination law by not interfering with the capacity of judges to contextually interpret comparative elements and discriminatory thresholds on a case-by-case basis. Rather, the measure provides the necessary statistical evidence to compare the magnitude of outcomes and potential disparity between all affected protected groups (for which data is available in a given case). It is thus a tool that enables identification and assessment of potential discrimination, but does not aim to make normative, fundamentally political case-specific determinations normally reserved for judicial interpretation, such as who is a legitimate comparator group or what is an appropriate threshold for illegal disparity in a given case. Adopting CDD as an evidential baseline for algorithmic discrimination would not interfere with the power of judges to interpret the facts of a case and make such crucial determinations.

As the everyday systems we interact with become more complex and significantly impact our lives, the limitations of our intuitive understanding of fairness and discrimination when applied to algorithms quickly become apparent. New tools to understand, prevent, and fix discrimination are needed not just to future proof legal remedies, but also to ensure close alignment between technological development and the fundamental rights and freedoms on which our societies are built. Making AI fair is a technically, legally, and socially complex challenge. We propose the use of conditional demographic disparity as a first step to help align technology and the law to tackle automated discrimination in Europe.[305]

---

[305] Of course, CDD suffers from the same requirements and limitations as many other fairness metrics and types of statistical evidence consisted in non-discrimination cases. For CDD to be effective data regarding protected attributes must be available or inferred. In calling for CDD as a baseline evidential standard we are thus implicitly calling for some protected attribute data to be collected and held for the purposes of statistical assessment. This is not a neutral request by any means and ties in with a broader societal and regulatory discussion around the need to share protected attributes to detect and prevent discrimination more consistently.

## APPENDIX 1 – MATHEMATICAL PROOF FOR CONDITIONAL DEMOGRAPHIC PARITY

This section contains a simple proof that (conditional) demographic parity implies the same proportion of people from each group are advantaged in each group, once you condition on the selected factors. Acknowledging its simplicity, it is included solely for the sake of completeness.

We consider a particular group corresponding to the set of attributes $R$. Returning to our hypothetical example of Company B (see: Section V.A), we write $B_A$ for the number of advantaged black people; $W_A$ for the number of advantaged white people and $B_D$ and $W_D$ for the numbers of disadvantaged white and black people respectively, then:

Starting from the definition of conditional demographic parity. For any choice of $R$ we have:

$$\frac{B_A}{B_A + W_A} = \frac{B_D}{B_D + W_D}$$

Assuming that neither of $B_A$ and $B_D$ are zero, we invert both sides to get:

$$\frac{B_A + W_A}{B_A} = \frac{B_D + W_D}{B_D}$$

Which is the same as:

$$\frac{B_A}{B_A} + \frac{W_A}{B_A} = \frac{B_D}{B_D} + \frac{W_D}{B_D}$$

$$1 + \frac{W_A}{B_A} = 1 + \frac{W_D}{B_D}$$

And therefore:

$$\frac{W_A}{B_A} = \frac{W_D}{B_D}$$

$$W_A B_D = W_D B_A$$

$$\frac{B_D}{B_A} = \frac{W_D}{W_A}$$

$$1 + \frac{B_D}{B_A} = 1 + \frac{W_D}{W_A}$$

$$\frac{B_D + B_A}{B_A} = \frac{W_D + W_A}{W_A}$$

Inverting again we arrive at:

$$\frac{B_A}{B_D + B_A} = \frac{W_A}{W_D + W_A}$$

Or that the proportion of advantaged black people is the same as the proportion of advantaged white people, and we can write this proportion as $k_R$% as required.